\documentclass[10pt,twocolumn,letterpaper]{article}

\usepackage[pagenumbers]{cvpr} 

\usepackage{graphicx}
\usepackage{amsmath}
\usepackage{amssymb}
\usepackage{booktabs}
\usepackage[table]{xcolor}
\usepackage{multirow}
\usepackage{floatrow}
\usepackage{makecell}
\usepackage{enumitem}
\usepackage{pifont}
\setlist{nosep}

\usepackage[symbol]{footmisc}

% ...
\usepackage[pagebackref,breaklinks,colorlinks]{hyperref}

\usepackage[capitalize]{cleveref}
\crefname{section}{Sec.}{Secs.}
\Crefname{section}{Section}{Sections}
\Crefname{table}{Table}{Tables}
\crefname{table}{Tab.}{Tabs.}

\def\cvprPaperID{4752} 
\def\confName{CVPR}
\def\confYear{2023}

\begin{document}

\title{UniHCP: A Unified Model for Human-Centric Perceptions} 

\author{Yuanzheng Ci\textsuperscript{1}\thanks{Equal contribution.}, Yizhou Wang\textsuperscript{2,3}\footnotemark[\value{footnote}], Meilin Chen\textsuperscript{2}, Shixiang Tang\textsuperscript{1}, Lei Bai\textsuperscript{3}\thanks{Corresponding author.},
Feng Zhu\textsuperscript{4}, Rui Zhao\textsuperscript{4,5}, \\Fengwei Yu\textsuperscript{4},
Donglian Qi\textsuperscript{2},
Wanli Ouyang\textsuperscript{3} \\
\textsuperscript{1}The University of Sydney,  \textsuperscript{2}Zhejiang University,  \textsuperscript{3}Shanghai AI Laboratory,  \textsuperscript{4}SenseTime Research,\\ \textsuperscript{5}Qing Yuan Research Institute, Shanghai Jiao Tong University, Shanghai, China \\
\tt\small yuanzheng.ci@sydney.edu.au, yizhouwang@zju.edu.cn, bailei@pjlab.org.cn
}

\maketitle

\begin{abstract}
    Human-centric perceptions (e.g., pose estimation, human parsing, pedestrian detection, person re-identification, etc.) play a key role in industrial applications of visual models. While specific human-centric tasks have their own relevant semantic aspect to focus on, they also share the same underlying semantic structure of the human body.  However, few works have attempted to exploit such homogeneity and design a general-propose model for human-centric tasks. 
    In this work, we revisit a broad range of human-centric tasks and unify them in a minimalist manner. We propose UniHCP, a \textbf{Uni}fied Model for \textbf{H}uman-\textbf{C}entric 
    \textbf{P}erceptions, which unifies a wide range of human-centric tasks in a simplified end-to-end manner with the plain vision transformer architecture. 
    With large-scale joint training on 33 human-centric datasets, 
    UniHCP can outperform strong baselines on several in-domain and downstream tasks by direct evaluation.
    When adapted to a specific task, UniHCP achieves new SOTAs on a wide range of human-centric tasks, e.g., 69.8 mIoU on CIHP for human parsing, 86.18 mA on PA-100K for attribute prediction, 90.3 mAP on Market1501 for ReID, and 85.8 JI on CrowdHuman for pedestrian detection,  performing better than specialized models tailored for each task. The code and pretrained model are available at \href{https://github.com/OpenGVLab/UniHCP}{https://github.com/OpenGVLab/UniHCP}.

\end{abstract}
\section{Introduction}
\label{sec:intro}

\begin{figure}[t]
\centering
\includegraphics[height=5.3cm]{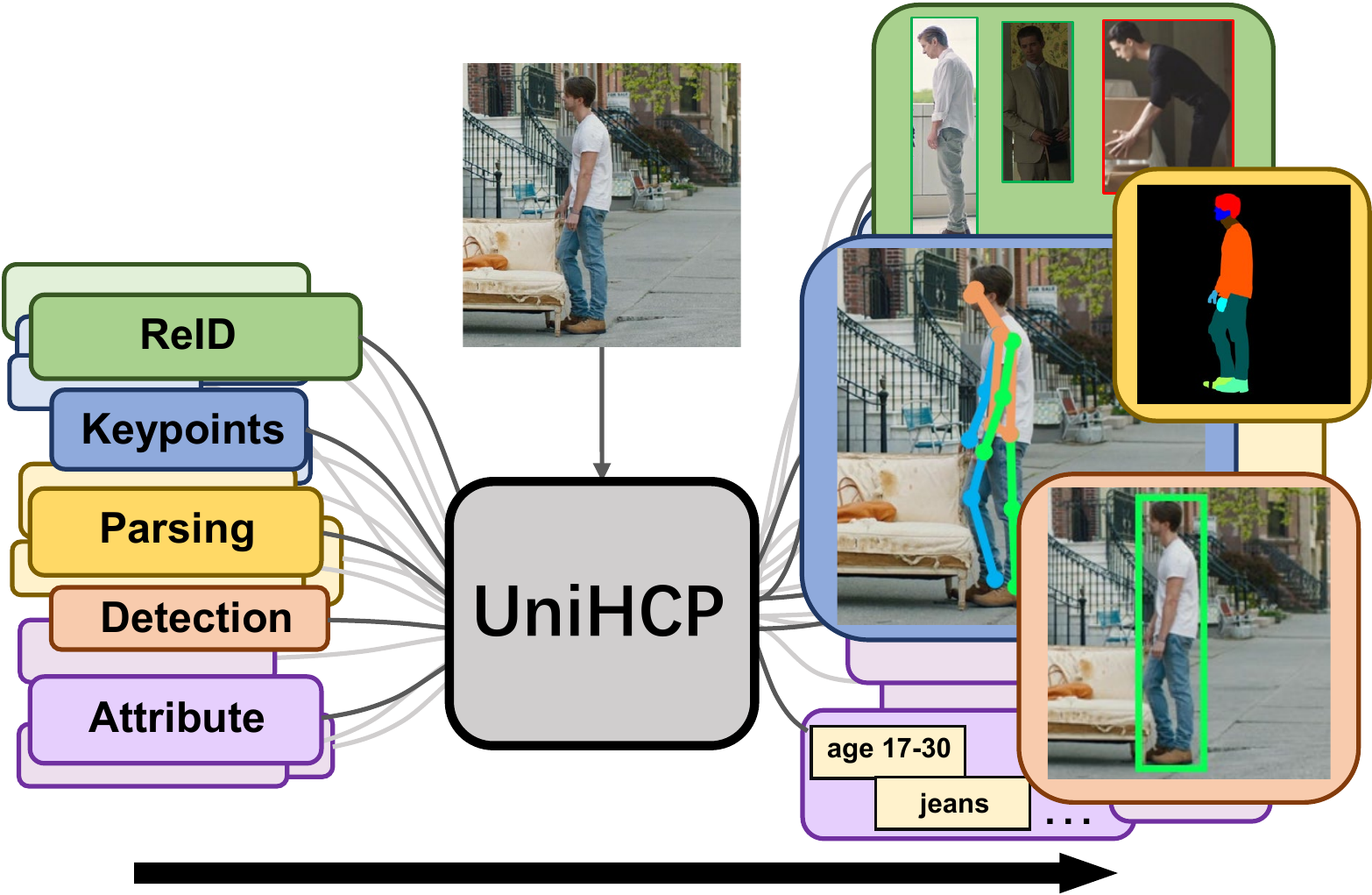}
\caption{UniHCP unifies 5 human-centric tasks under one model and is trained on a massive collection of human-centric datasets.}
\label{fig:intro}
\end{figure}
Research on human-centric perceptions has come a long way with tremendous advancements in recent years. Many methods have been developed to enhance the performance of pose estimation~\cite{munea2020progress}, human parsing~\cite{liang2018look}, pedestrian detection~\cite{cao2021handcrafted}, and many other human-centered tasks. These significant progress play a key role in advancing the applications of visual models in numerous fields, such as sports analysis~\cite{cust2019machine}, autonomous driving~\cite{zhang2019widerperson}, and electronic retailing~\cite{kalantidis2013getting}.

Although different human-centric perception tasks have their own relevant semantic information to focus on, those semantics all rely on the same basic structure of the human body and the attributes of each body part~\cite{park2017attribute, wang2020hierarchical}. 
In light of this, there have been some attempts trying to exploit such homogeneity and train a shared neural network jointly with distinct human-centric tasks~\cite{su2017multi, khamis2014joint, liang2018look, tian2015pedestrian, zhang2019pose2seg, nie2018mutual,xu2022fashionformer,lin2018multi,kalayeh2018human}. For instance, human parsing has been trained in conjunction with human keypoint detection~\cite{liang2018look, zhang2019pose2seg, nie2018mutual}, pedestrian attribute recognition~\cite{xu2022fashionformer}, pedestrian detection~\cite{lin2018multi} or person re-identification~\cite{kalayeh2018human} (ReID).
The experimental results of these works empirically validate that some human-centric tasks may benefit each other when trained together.
Motivated by these works, a natural expectation is that a more versatile all-in-one model could be a feasible solution for general human-centric perceptions, which can utilize the homogeneity of human-centric tasks for improving performance, enable fast adaption to new tasks, and decrease the burden of memory cost in large-scale multitask system deployment compared with specific models to specific tasks.

However, unifying distinct human-centric tasks into a general model is challenging considering the data diversity and output structures. From the data's perspective, images in different human-centric tasks and  different datasets have different resolutions and characteristics (e.g., day and night, indoor and outdoor), which calls for a robust representative network with the capability to accommodate them. From the perspective of output, the annotations and expected outputs of different human-centric tasks have distinct structures and granularities. 
Although this challenge can be bypassed via deploying separate output heads for each task/dataset, it is not scalable when the number of tasks and datasets is large.

In this work, we aim to explore a simple, scalable formulation for unified human-centric system and, for the first time, propose a \underline{Uni}fied model for \underline{H}uman-\underline{C}entric \underline{P}erceptions (UniHCP). As shown in Figure.\ref{fig:intro}, UniHCP unifies and simultaneously handles five distinct human-centric tasks, namely, pose estimation, semantic part segmentation, pedestrian detection, ReID, and person attribute recognition. 
Motivated by the extraordinary capacity and flexibility of the vision transformers~\cite{li2022exploring,zhai2022scaling}, a simple yet unified encoder-decoder architecture with the plain vision transformer is employed to handle the input diversity, which works in a simple feedforward and end-to-end manner, and can be shared across all human-centric tasks and datasets to extract general human-centric knowledge. 
To generate the output for different tasks with the unified model, UniHCP defines Task-specific Queries, which are shared among all datasets with the same task definition and interpreted into different output units through a Task-guided Interpreter shared across different datasets and tasks.
With task-specific queries and the versatile interpreter, UniHCP avoids the widely used task-specific output heads, which minimizes task-specific parameters for knowledge sharing and make backbone-encoded features reusable across tasks.

Own to these designs, UniHCP is suitable and easy to perform multitask pretraining at scale. To this end, we pretrained an UniHCP model on a massive collection of 33 labeled human-centric datasets. By harnessing the abundant supervision signals of each task, we show such a model can simultaneously handle these in-pretrain tasks well with competitive performance compared to strong baselines relying on specialized architectures. When adapted to a specific task, both in-domain and downstream, our model achieves new SOTAs on several human-centric task benchmarks. 
In summary, the proposed model has the following properties:

\begin{itemize}[noitemsep]
  \item[\tiny\ding{110}] Unifying five distinct human-centric tasks and handling them simultaneously.
  \item[\tiny\ding{110}] Shared encoder-decoder network based on plain transformer.
  \item[\tiny\ding{110}] Simple task-specific queries identifying the outputs. 
  \item[\tiny\ding{110}] Maximum weight sharing (99.97\% shared parameters) with a task-guided interpreter.
  \item[\tiny\ding{110}] Trainable at scale and demonstrates competitive performance compared to task-specialized models.
\end{itemize} 

The following sections are organized as follows: Section~\ref{sec:related} reviews the related works with focuses on Human-Centric perception and unified models. Section~\ref{sec:unihcp} describes the proposed model. Section~\ref{sec:exps} provides implementation details together with empirical results and ablation studies. Finally, we conclude the paper in Section~\ref{sec:conclu}.

\section{Related Works}
\label{sec:related}

\subsection{Human-Centric Perceptions}
Human-centric perceptions are essential for substantial real-world applications. Depending on the targeted visual concept, the way of decoding output from image features varies across tasks. Specifically, pose estimation and pedestrian detection are both localization tasks that can be solved by either regression-based methods~\cite{li2021human,zhang2016far} or heatmap-based methods~\cite{xiao2018simple,law2018cornernet,law2019cornernet}. Human parsing, as a fine-grained segmentation problem, is usually solved by per-pixel classification. While contour-based methods~\cite{xie2020polarmask,peng2020deep} can also obtain segmentation masks, it requires instance-level mask annotations, which are not always available. PAR is treated as a multi-label classification task~\cite{zhu2017multi}, and ReID is treated as a feature learning task~\cite{sun2020circle}.

Recently, several transformer-based solutions have been proposed for these human-centric tasks, with attention block designs on both backbone~\cite{yuan2021hrformer,he2021transreid,xu2022vitpose} and decoding network~\cite{li2021tokenpose,mao2022poseur,liu2021query2label,li2022label2label,xu2022fashionformer,zheng2022progressive}. However, these methods involve \textit{different} task-specific designs and thus cannot be integrated into one model seamlessly. Built upon the general success of these works, we take a further step and unify human-centric tasks under the \textit{same} architecture based on plain vision transformer.


\begin{figure*}[t]
    \centering
    \includegraphics[height=6.5cm]{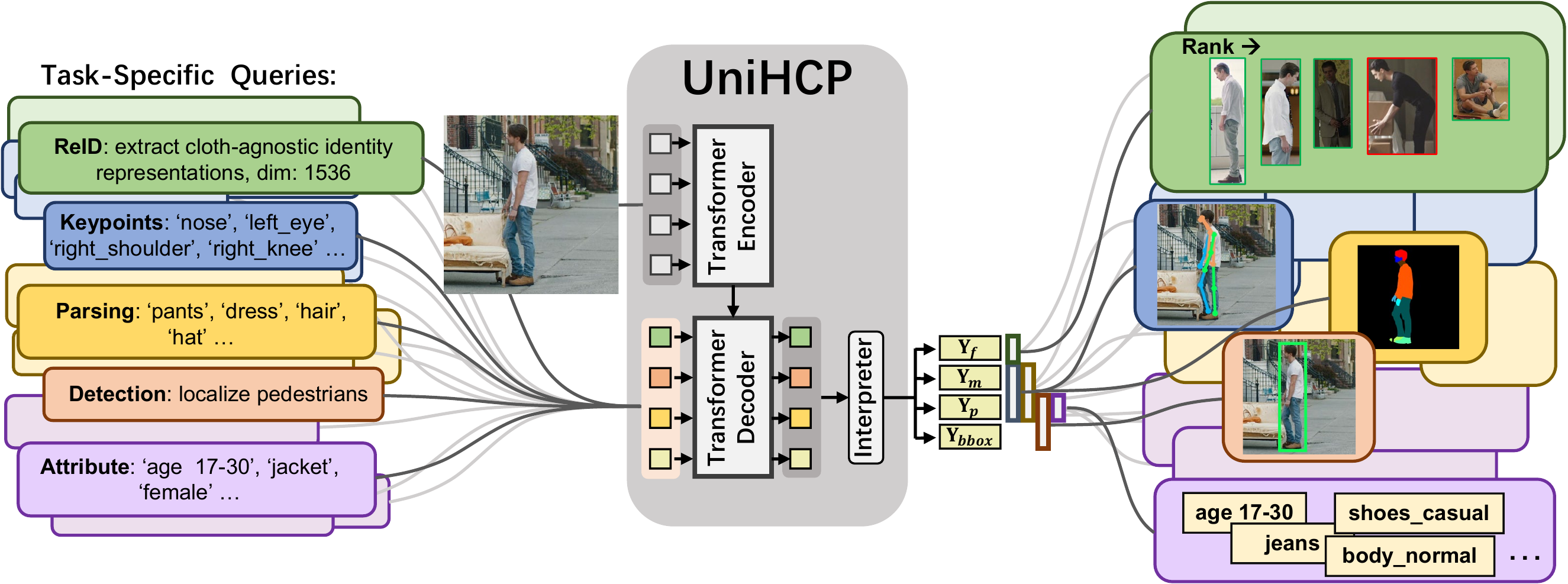}
    \caption{UniHCP handles a massive collection of human-centric tasks  uniformly by task-specific queries and a task-guided interpreter, all predictions are yielded in parallel through a simple encoder-decoder transformer architecture.}
    \label{fig:arch}
\end{figure*}

\subsection{Unified Models}
A general-purpose model that can handle different tasks in a unified manner has long been a coveted alternative to models specifically tailored for different tasks.
Pioneering works regarding Natural Language Processing (NLP)~\cite{raffel2020exploring}, vision-language~\cite{lu202012}, and basic vision tasks~\cite{sermanet2013overfeat,kokkinos2017ubernet} have shown the effectiveness of such kind of unified cross-task models. 
ExT5~\cite{aribandi2021ext5} and OFA~\cite{wang2022ofa} further provide a degree of promise for the performance benefits of large-scale multitask co-training.
Among models supporting visual tasks, UniHead~\cite{liang2022unifying} and UViM~\cite{kolesnikov2022uvim} propose a unified architecture for several vision tasks. However, they are only trained and evaluated in a single-task manner.

For methods supporting multitask co-training, Uni-Perceiver~\cite{zhu2022uni} focuses on tasks in which the desired output is inherently language or labels, which does not fit human-centric tasks. While UniT~\cite{hu2021unit}, OFA~\cite{wang2022ofa}, Unified-IO~\cite{lu2022unified}, and Pix2Seq v2~\cite{chen2022unified} further extend the support for detection, keypoint detection, segmentation, and many other visual tasks, they rely on \emph{independent decoder heads}~\cite{hu2021unit,wang2022ofa} or \emph{autoregressive} modeling~\cite{chen2022unified,lu2022unified}. These works do not focus on  human-centric vision tasks.
Differently, our work introduces a \emph{shared decoder head} (task-guided interpreter) in a \emph{parallelly feedforward} manner for human-centric vision tasks, which is simple yet maximizes the parameter sharing among different tasks. 

In the case of human-centric tasks, many works have shown great success by co-training a pair of human-centric tasks~\cite{su2017multi, khamis2014joint, liang2018look, tian2015pedestrian, zhang2019pose2seg, nie2018mutual,xu2022fashionformer,lin2018multi,kalayeh2018human}. However, there is no work exploring a general unified model that can handle all representative human-centric tasks. Our work is the first attempt at designing, training, and evaluating a unified human-centric model with a large-scale multitask setting.






\section{UniHCP}
\label{sec:unihcp}
\begin{table}[t]
  \centering
  \footnotesize
  \caption{Network details of UniHCP}
    \begin{tabular}{lccc}
    \toprule
          & \multicolumn{1}{l}{Layers} & \multicolumn{1}{l}{Dimension} & Params \\
    \midrule
    Encoder & 12    & 768   & 91.1M \\
    Decoder & 9     & 256   & 14.5M \\
    Task-guided Interpreter &       &       & 3.5M \\
    \midrule
    Task-specific queries &       & 256   & \textless{}0.03M \\
    \midrule
    Total  &       &       & 109.1M \\
    \multicolumn{3}{l}{Task-agnostic params / total params}        & 99.97\% \\
    \bottomrule
    \end{tabular}%
  \label{tab:addlabel}%
\end{table}%
To share the most knowledge among various human-centric tasks, we attempt to maximize weight sharing among all tasks in UniHCP. Specifically, our UniHCP, as shown in Figure~\ref{fig:arch}, consists of three components: (1) A task-agnostic transformer encoder $E$ to extract image features. (2) A transformer decoder $D$ that attends to task-specific information according to task-specific queries $\{\mathbf{Q}^t\}$, where $t$ denotes a specific task. (3) A task-guided interpreter $\mathcal{I}$ produces output units, in which we decompose the output of multiple human-centric perception tasks into sharable units of diverse granularities, \emph{i.e.,} feature representation, local probability map, global probability, bounding box coordinates. Since only the queries to the decoders are not shared among tasks, we can learn human-centric knowledge across different granularities by the designed interpreters and achieve maximum parameter sharing among all tasks, i.e., \underline{\textbf{99.97\%}} shared parameters, as shown in Table~\ref{tab:addlabel}. The pipeline for our UniHCP is described as follows.

\emph{Step 1}: Given an image $\mathbf{X}$ sampled from the dataset in  task $t$, extract encoded features $\mathbf{F}$ by the task-agnostic transformer encoder $E$ (Sec.~\ref{sec:encoder}).

\emph{Step 2} : A transformer decoder $D$ with task-specific queries $\mathbf{Q}^t$ extracts task-specific features from encoded features $\mathbf{F}$ (Sec.~\ref{sec:decoder}). 

\emph{Step 3}: Generate output units according to the queried task, i.e., attended features $\mathbf{Y}_f$, local probability map $\mathbf{Y}_m$, global probability $\mathbf{Y}_p$ and bounding box coordinates $\mathbf{Y}_{bbox}$ by a task-guided interpreter $\mathcal{I}$ (Sec.~\ref{sec:interpreter}). For example, for human parsing, two units: local probability map $\mathbf{Y}_m$ (for semantic part segmentation) and global probability $\mathbf{Y}_p$ (for existence of body part in the image), are generated.

\emph{Step 4:} Calculate the loss of the corresponding task for optimizing the encoder $E$, the decoder $D$, the task-specific queries $\mathbf{Q}^t$ and task-guided interpreter $\mathcal{I}$ by backward propagation (Sec.~\ref{sec:objectives}).

\subsection{Task-agnostic Transformer Encoder} \label{sec:encoder}
UniHCP uses a plain Vision Trasnformer~\cite{dosovitskiy2020image} (ViT) as the encoder. To handle input images of different resolutions, we use a shared learnable positional embedding with the size of $84 \times 84$ and interpolate it based on the spatial size of the input image after patch projection. The encoded feature $\mathbf{F}$ can be mathematically calculated as
\begin{equation}
    \mathbf{F} = E(\mathbf{X}, \mathbf{P}_E),
\end{equation}
where $\mathbf{P}_E$ is the positional embedding after interpolation and $E$ denotes the task-agnostic transformer encoder.

\subsection{Decoder with Task-specific Queries} \label{sec:decoder}
To obtain the most discriminative feature for each task while maximizing knowledge sharing, we design task-specific queries to guide the  transformer decoder only attending to task-relevant information. 

\noindent \textbf{Task-specific Queries.}
Task queries for task $t$ are denoted as
\begin{equation} \label{eq:task_queries}
    \mathbf{Q}^t=[\mathbf{q}^t_1, \mathbf{q}^t_2, ..., \mathbf{q}^t_{N^t}],
\end{equation}
where $N^t$ denotes the number of queries representing $N^t$ different semantic meanings in task $t$. For pedestrian attribute recognition, pose estimation, human parsing, and ReID, the number of queries respectively equals to the number of  attributes, the number of pose joints, the number of semantic parsing classes, and the length of desired ReID features. For pedestrian detection, we follow the implementation in~\cite{wang2022anchor}, with details provided in the supplementary material. 
We randomly initialize the task-specific query $\mathbf{Q}^t$ as learnable embeddings $\mathbf{Q}_0^t$ and refine it with the following decoder blocks. 


Following the common practice as in~\cite{vaswani2017attention,cheng2022masked,wang2022anchor}, all $\mathbf{Q}^t$ are also associated with a positional embedding $\mathbf{Q}^{t}_{p}$ 
, which has the same dimension as $\mathbf{Q}^t$ and is not shared across tasks. Different from $\mathbf{Q}^t$ that will be progressively refined in the decoder blocks, $\mathbf{Q}^{t}_{p}$ is shared across decoder blocks. 
For tasks other than pedestrian detection, $\mathbf{Q}^{t}_{p}$ is simply a learnable positional embedding that is randomly initialized.
For pedestrian detection, we have
\begin{equation}
\mathbf{Q}^{t}_{p} = proj(\mathcal{A}_{\mathbf{Q}}),
\end{equation}
where $\mathcal{A}_{\mathbf{Q}} \in \mathbb{R}^{N^t \times 2}$ refers to $N^t$ learnable anchor points that are initialized with a uniform distribution following~\cite{wang2022anchor}, and $proj$ is a projection from coordinates to positional embeddings (more details about the projector are elaborated in the supplementary materials).

\noindent\textbf{Decoder.} The transformer decoder aims to attend to task-specific features according to the task queries. We follow the standard design of transformer decoders~\cite{vaswani2017attention}. In the decoder, each transformer block $D_l$ for $l=1,2,...,L$ consists of a cross-attention module, a self-attention module, and a feed-forward module (FFN), where $L$ denotes the number of transformer blocks. We place cross-attention before self-attention as adopted by~\cite{cheng2022masked,li2022mask}. For each block $D_l$, we attend to task-specific information from the encoded feature by task queries, which can be 
formulated as 
\begin{equation}
\label{eq:decoder}
    \mathbf{Q}_l^t =  D_l(\mathbf{Q}^t_{l-1}, \mathbf{Q}^t_{p}, \mathbf{F}, \mathbf{F}_{p}),
\end{equation}
\begin{equation}
\textrm{where }\mathbf{F}_{p} = proj(\mathcal{A}_{\mathbf{\mathbf{F}}}),
\end{equation}
$\mathcal{A}_{\mathbf{\mathbf{F}}} \in \mathbb{R}^{H_{\mathbf{F}}W_{\mathbf{F}} \times 2}$ is the coordinates with respect to the original image for all feature tokens in $\mathbf{F} \in R^{H_{\mathbf{F}} \times W_{\mathbf{F}}}$. For the cross-attention in the decoder $D_l$, the query is $\hat{\mathbf{Q}}^t_l = \mathbf{Q}^t_{l-1}+\mathbf{Q}^t_{p}$, the key is $\hat{\mathbf{K}} = \mathbf{F'} + \mathbf{F}_{p}$, and the value is $\hat{\mathbf{V}} = \mathbf{F'}$, where $\mathbf{F}'$ is linearly projected from the features of the encoder $\mathbf{F}$ to align channel dimensions.
The result of cross-attention is then passed for self-attention in $D_l$.

\begin{figure}[t]
\centering
\includegraphics[height=5.0cm]{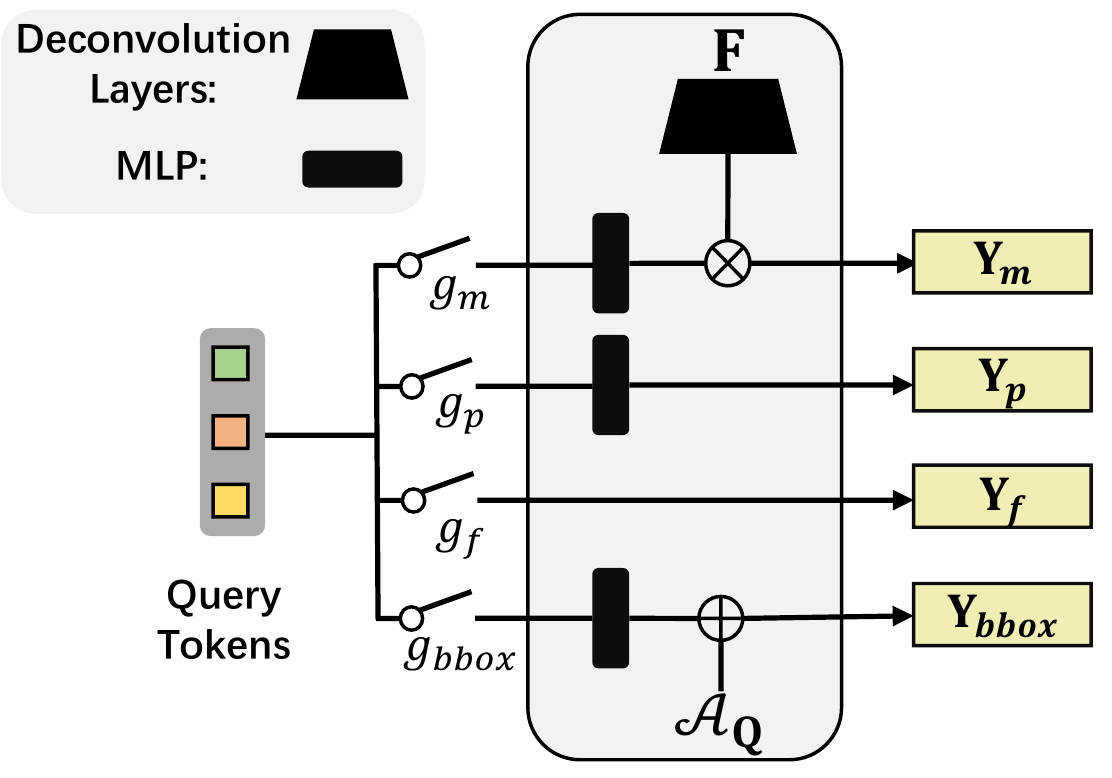}
\caption{Task-guided interpreter.  $\otimes$ denotes a dynamic convolution module~\cite{chen2020dynamic} that takes the projected query feature as the kernel and takes the tokens $\mathbf{F}$ from the encoder as the feature map, where $\mathbf{F}$ is upscaled to the desired resolution $H' \times W'$ , $\oplus$ denotes addition, for which the inputs are the projected query feature in the format of $[\nabla cx, \nabla cx, h, w]$ and $\mathcal{A}_{\mathbf{Q}}$, which contains the anchor point $[cx, cy]$ (see supplementary materials for details). 
}
\label{fig:interp}
\end{figure}
\subsection{Task-guided Interpreter} \label{sec:interpreter}
Task-guided interpreter $\mathcal{I}$ interprets  query tokens $\mathbf{Q}^t$ into 
four output units subject to the request of a specific task.
As shown in Figure~\ref{fig:interp}, these four output units are as follows: 

\begin{equation} \label{eq:activation}
\begin{aligned}
\text{feature vector unit}:&\  \mathbf{Y}_f \in \mathbb{R}^{N^t \times C}\\
\text{global probability unit}:&\  \mathbf{Y}_p \in \mathbb{R}^{N^t \times 1}\\
\text{local probability map unit}:&\  \mathbf{Y}_m \in \mathbb{R}^{N^t \times H' \times W'}\\
\text{bounding box unit}:&\ \mathbf{Y}_{bbox} \in \mathbb{R}^{N^t \times4},
\end{aligned}
\end{equation}
where $C$ is the output dimension of the decoder, $H' \times W'$ denotes the desired resolution for the local probability map.
Given task $t$ and output interpreter $\mathcal{I}$, the output of the UniHCP is defined as follows:
\begin{equation} \label{eq:gated}
\begin{aligned}
\{\mathbf{Y}_u|g_{u}^{\mathbf{t}_t} = 1, u \in \{f,p,m,bbox\}\} = \mathcal{I}(\mathbf{Q}^t, \mathbf{g}^{\mathbf{t}_t}),
\end{aligned}
\end{equation}
where $ \mathbf{t}_t \in \{reid, \ldots,pose\}$ denotes the task type of task $t$, $\mathbf{g}^{\mathbf{t}} = \{g^\mathbf{t}_u\}$ is a set of task-specific binary gates ($g \in \{0, 1\}$) that represents the desired output units for task type $\mathbf{t}$. 

\noindent\textbf{Guidance from tasks to output units.} For human parsing, local probability map (for semantic part segmentation) and global probability (for existence
of body part in the image) are activated, corresponding to $g^{seg}_m=1$ and $g^{seg}_p=1$ respectively. For
person ReID, feature vectors are used, corresponding to $g^{reid}_f=1$. For pose estimation, $g^{pose}_m=1$ (for localizing key points) and $g^{pose}_p=1$ (for existence of keypoints in the image). For detection, $g^{det}_{bbox}=1$ (for bounding box prediction) and $g^{det}_p=1$ (for existence of object). For pedestrian attribute prediction, $g^{par}_p=1$ (for existence of attributes in the image). Therefore, the output unit of global probabilities is shared among pose estimation, human parsing, pedestrian detection, and attribute recognition. The output unit of local probability maps is shared among pose estimation and human parsing.

\noindent\textbf{Discussion.} The task-guided interpreter interprets each query token independently. Previous works focused on autoregressive decoding with tokenization~\cite{chen2022unified, lu2022unified} or task-specific heads~\cite{yuan2021florence,hu2021unit} to handle different output units required by specific tasks. In contrast, the task-guided interpreter can handle tasks involving a varying number of classes, yield all results in parallel, and do not require task-specific heads. 
This is achieved by two designs in our UniHCP framework: 1) Class/instance information is self-contained in queries. As mentioned in Section~\ref{sec:decoder}, a query represents a particular semantic class in pose estimation, attribute prediction, human parsing, and pedestrian detection. We only need to retrieve a scalar probability value from a query to obtain the confidence information for a particular class/human instance. 2) Outputs of the same modality share the same output unit. For example, the heatmap for a particular joint in pose estimation and the heatmap for a particular body part in human parsing have the same dimension. 
Although these outputs have different meanings, experimental results in Section~\ref{sec:ablation} show that it is suitable to obtain them through the same output unit and fully let the task-specific queries handle the differences in preferred information to be represented.

\subsection{Objective Functions} \label{sec:objectives}
In this section, we will introduce the objective functions for training diverse human-centric tasks together and illustrate how these objectives are related to the output units defined in Eq.~\ref{eq:activation}. 
Unless otherwise specified, we omit the GT inputs in loss functions for brevity.

\noindent\textbf{Overall Objective Function.}
Given a collection of datasets $\mathbb{D} = \{\mathcal{D}| \mathbf{t}_{\mathcal{D}} \in \{reid, \ldots,pose\}\}$, where $\mathbf{t}_{\mathcal{D}}$ denotes the task type of dataset $\mathcal{D}$, we also note $t_{\mathcal{D}}$ as the task of dataset $\mathcal{D}$, we have the overall loss defined as:
\begin{equation}
 \mathcal{L} = \sum_{\mathcal{D} \in \mathbb{D}}w_D\mathcal{L}_{\mathbf{t}_\mathcal{D}}(\mathcal{I}(\mathbf{Q}^{t_{\mathcal{D}}}, \mathbf{g}^{\mathbf{t}_{\mathcal{D}}})),
\end{equation}
where $w_\mathcal{D}$ is the loss weight for dataset $\mathcal{D}$, which is calculated based on the task type and batch size (calculations are elaborated in supplementary materials).

\paragraph{ReID.}
Person ReID is a feature learning task for extracting identification information. Therefore, we directly supervised the features after the decoder by identity annotations. Specifically, for ReID task, the extracted feature is a simple concatenation of all feature vectors $\mathbf{Y}_f = [y_{f}^{1}; \ldots ;y_{f}^{N^t}]$, where $N^t = 6$ by default. The loss function is a combination of ID loss~\cite{zheng2017discriminatively} and triplet loss~\cite{liu2017end} written as follows:

\begin{equation}
\begin{aligned}
\mathcal{L}_{reid} = \mathcal{L}_{ID}(\mathbf{Y}_f) + \mathcal{L}_{triplet}(\mathbf{Y}_f).
\end{aligned}
\end{equation}


\paragraph{PAR.}
Pedestrian attribute recognition only predicts whether an attribute exists in the global image. Therefore, we only supervise the output unit of global probabilities $\mathbf{Y}_p$ from the task-guided interpreter. Specifically, following the common practice~\cite{tang2019improving,li2022label2label}, we adopt the weighted binary cross-entropy loss. Given the probability predictions $\mathbf{Y}_p$ associated with $N^t$ attributes, we have:
\begin{equation}
\begin{aligned}
\mathcal{L}_{par} &= \sum_{n=1}^{N_{t}}w_{n}(y_{n}\log(y_{p}^{n}) + (1-y_n)\log(1-y_{p}^{n})),\\
w_{n} &= y_{n}e^{1-\gamma_{n}} + (1-y_{n})e^{\gamma_{n}},
\end{aligned}
\end{equation}
where $y_n$ denotes the annotation of $n$-th attribute and $\gamma_{n}$ denotes the positive example ratio of $n$-th attribute. 

\paragraph{Human Parsing.}
Human parsing can be considered as semantic segmentation of human part. We view the presence of semantic classes as predictable attributes since the semantic classes are not always present in an image. Therefore, the global probability $\mathbf{Y}_p$ and local probability map $\mathbf{Y}_m$ are selected from the output units to describe whether a semantic part exists on the image level (global) and pixel level (local), respectively. Given a query $\mathbf{q}_l$ defined in Eq.~\ref{eq:task_queries} which corresponds to a semantic class in human parsing, we adopt the binary cross entropy loss as $\mathcal{L}_{par}$ in pedestrian attribute recognition to constrain the global probability $\mathbf{Y}_p$, and a combination of binary cross-entropy loss and dice loss~\cite{cheng2022masked} to supervised local probability map $\mathbf{Y}_m$ as follows:
\begin{equation}
\begin{aligned}
\mathcal{L}_{seg} = \lambda_{par}\mathcal{L}_{par}(\mathbf{Y}_p) + \mathcal{L}_{bce}(\mathbf{Y}_m) + \mathcal{L}_{dice}(\mathbf{Y}_m), 
\end{aligned} \notag
\end{equation}
where $\lambda_{par}$ denotes the loss weight for $\mathcal{L}_{par}(\mathbf{Y}_p)$.

\paragraph{Pose Estimation.}
We follow the common top-down setting for pose estimation, i.e., predicting keypoints based on the cropped human instances. We predict the heatmap w.r.t. the keypoints via mean-squared error. Similar to human parsing formulation, we also select the global probability $\mathbf{Y}_p$  and local probability map $\mathbf{Y}_m$ to predict whether a keypoint exists in the image level and pixel level, respectively. Mathematically, we have:
\begin{equation}
\begin{aligned}
\mathcal{L}_{pose} = \lambda_{par}\mathcal{L}_{par}(\mathbf{Y}_p) + \mathcal{L}_{mse}(\mathbf{Y}_m).
\end{aligned}
\end{equation}

\paragraph{Pedestrian Detection.}
Pedestrian Detection is a local prediction task but in a sparse manner. 
Following the widely adopted designs in end-to-end transformer-based detection~\cite{carion2020end,zheng2022progressive}, ground-truth for $N^t$ query features in $\mathbf{Q}_l$ are determined by optimal bipartite matching between all $N^t$ predictions and GT boxes. Given output units $\mathbf{Y}_{p}$ and $\mathbf{Y}_{bbox}$, we adopt the identical cost formulation and loss as in~\cite{zheng2022progressive},
\begin{equation}
\begin{aligned}
\mathcal{L}_{peddet} = &\lambda_{cls}\mathcal{L}_{cls}(\mathbf{Y}_p) + \lambda_{iou}\mathcal{L}_{iou}(\mathbf{Y}_{bbox}) + \\&\lambda_{L1}\mathcal{L}_{L1}(\mathbf{Y}_{bbox}).
\end{aligned}
\end{equation}
where  $\mathcal{L}_{cls}, \mathcal{L}_{iou}$ and $\mathcal{L}_{L1}$ are focal loss~\cite{lin2017focal}, GIoU loss~\cite{rezatofighi2019generalized}, and $L1$ loss, respectively. Their corresponding loss weights $\lambda$ are also identically set as in~\cite{zheng2022progressive}.  


\section{Experiments}
\label{sec:exps}
\subsection{Implementation details}
\noindent{\textbf{Datasets.}}
To enable general human-centric perceptions, we pretrain the proposed UniHCP at scale on a massive and diverse collection of human-centric datasets. Specifically, the training splits of 33 publically available datasets are gathered to form the training set for UniHCP, including nine datasets for pose estimation and six datasets for ReID, Human Parsing, Attribute Prediction, Pedestrain Detection, seraprately. 
For ReID, there are two different sub-tasks: general ReID and cloth-changing ReID, where the difference is whether cloth-change is considered for person ReID. We empirically found it is best to view them as different tasks and solve them with different task queries. Hence, we treat these two sub-tasks as different tasks and give them separate queries. 

We carefully follow the de-duplication practices as introduced in~\cite{zhai2022lit} to remove the samples that could appear in the evaluation datasets. We also remove images whose groundtruth labels are not given, leading to 2.3M distinct training samples in total. 
For evaluation, apart from the available validation or test splits of the 33 training sets, we also included several out-of-pretrain downstream datasets for each type of human-centric task. 
More details about dataset setups can be found in supplementary materials.
\begin{table}[]
\footnotesize
  \centering
  \caption{Representative datasets used in multitask co-training.}
\begin{tabular}{llc}
\toprule
Task Type                                                                                    & Datasets        & Number of samples\\ \hline
\multirow{4}{*}{\begin{tabular}[c]{@{}l@{}}ReID\\ (6 datasets)\end{tabular}}                 & CUHK03~\cite{li2014deepreid}    & \multirow{4}{*}{268,002}     \\ 
                                                                                             & DGMarket~\cite{zheng2019joint}       \\ 
                                                                                             & PRCC~\cite{yang2019person}            \\
                                                                                             & ...             \\ \midrule
\multirow{4}{*}{\begin{tabular}[c]{@{}l@{}}Pose Estimation\\ (9 datasets)\end{tabular}}      & COCO-Pose~\cite{lin2014microsoft} & \multirow{4}{*}{1,261,749}      \\
                                                                                             & AI Challenger~\cite{wu2019large}  &   \\
                                                                                             & PoseTrack~\cite{andriluka2018posetrack}       \\
                                                                                             & ...             \\ \midrule
\multirow{4}{*}{\begin{tabular}[c]{@{}l@{}}Human Parsing\\ (6 datasets)\end{tabular}}        & LIP~\cite{gong2017look}      & \multirow{4}{*}{384,085}      \\
                                                                                             & CIHP~\cite{gong2018instance}   &          \\
                                                                                             & DeepFashion2~\cite{ge2019deepfashion2}    \\
                                                                                             & ...             \\ \midrule
\multirow{4}{*}{\begin{tabular}[c]{@{}l@{}}Attribute Prediction\\ (6 datasets)\end{tabular}} & PA-100K~\cite{liu2017hydraplus}  & \multirow{4}{*}{242,880}        \\
                                                                                             & RAPv2~\cite{li2019richly}   &         \\
                                                                                             & UAV-Human~\cite{li2021uav}        \\
                                                                                             & ...             \\ \midrule
\multirow{4}{*}{\begin{tabular}[c]{@{}l@{}}Pedestrian Detection\\ (6 datasets)\end{tabular}} & COCO-Person~\cite{lin2014microsoft}   & \multirow{4}{*}{170,687 }  \\
                                                                                             & CrowdHuman~\cite{shao2018crowdhuman} &     \\
                                                                                             & WiderPedestrian~\cite{loy2019wider} \\
                                                                                             & ...             \\ \bottomrule
\end{tabular}
\vspace{-1em}
\end{table}

\noindent{\textbf{Training.}}
We use the standard ViT-B~\cite{dosovitskiy2020image} as the encoder network and initialize it with the MAE pretrained~\cite{he2022masked} weights following~\cite{xu2022vitpose,li2022exploring}. For the main results, we use a batch size of 4324 in total, with the dataset-specific batch size being proportional to the size of each dataset. Unless otherwise specified, the image resolution used in pretraining is $256\times192$ for pose estimation and attribute prediction, $256\times128$ for ReID, $480\times480$ for human parsing, and a maximum height/width of 1333 for pedestrian detection. 

For computational efficiency, each GPU only runs one specific task, and each task can be evenly distributed to multiple GPUs whereas a single GPU is not capable of handling its workloads.
To further save the GPU memory during the training time, we adopt the gradient checkpointing~\cite{FairScale2021}
in the encoder forward pass among all tasks and additionally use accumulative gradients for detection tasks. Due to the high GPU-memory demand of detection datasets, the batch size for the detection task is timed by 0.6. 

We use Adafactor~\cite{shazeer2018adafactor} optimizer and follow the recommended modifications~\cite{zhai2022scaling} for adopting it to ViT, we set $\beta_1 = 0.9$, $\beta_2$ clipped at $0.999$, disables the parameter scaling and decoupled weight decay to 0.05. We linearly warm up the learning rate for the first 1500 iterations to 1e-3, after which the learning rate is decayed to 0 following a cosine decay scheduler.
We also use a drop-path rate of 0.2 and layer-wise learning rate decay~\cite{li2022exploring,xu2022vitpose} of 0.75 in the ViT-B encoder.
For the main results, the whole training process takes 105k iterations which are approximately 130 epochs for detection datasets and 200 epochs for other datasets. The whole training takes 120 hours in total on 88 NVIDIA V100 GPUs.


\begin{table*}[ht]
\begin{floatrow}
\floatbox{table}[0.755\linewidth]
{\caption{\small{Person ReID evaluation on Market1501, MSMT, CUHK03 with mAP. \dag indicates using additional camera IDs. }}
 \label{tab:reid} }
{
\footnotesize
\resizebox{\linewidth}{!}{
\begin{tabular}{lccc}
    \toprule
    Method & Market1501 & MSMT17  & CUHK03 \\
    \midrule
    HOReID~\cite{wang2020high} & 84.9  & -     & - \\
    MNE~\cite{li2019memory}& -     & -     & 77.7  \\
    SAN~\cite{jin2020semantics}   & 88.0  & 55.7  & 76.4  \\
    TransReID~\cite{he2021transreid} & 86.8  & 61.0  & - \\
    TransReID\dag~\cite{he2021transreid} & 88.9  & \textbf{67.4}  & - \\
    \midrule
    UniHCP (direct eval) & 80.7
  & 55.2  & 68.6 \\
    UniHCP (finetune) & \textbf{90.3}  & 67.3  & \textbf{83.1} \\
    \bottomrule
    \end{tabular}
    }
    }
\hspace{-1.58em}
\floatbox{figure}[.57\linewidth]
{\captionof{table}{\small{Pedestrian attribute recognition evaluation on PA-100K and RAPv2 test sets with mA. }}
 \label{tab:attribute}}
{ \footnotesize
\resizebox{\linewidth}{!}{
   \begin{tabular}{lcc}
    \toprule
    Method & PA-100K & RAPv2 \\
    \toprule
    SSC~\cite{jia2021spatial}   & 81.87  & - \\
    C-Tran~\cite{lanchantin2020general} & 81.53  & - \\
    Q2L~\cite{liu2021query2label}   & 80.72  & - \\
    L2L~\cite{li2022label2label}   & 82.37  & - \\
    DAFL~\cite{jia2022learning}  & 83.54  & 81.04  \\
    \midrule
    UniHCP (direct eval) & 79.32  & 77.20  \\
    UniHCP (finetune) & \textbf{86.18}  & \textbf{82.34}  \\
    \bottomrule
    \end{tabular}}
    }
\hspace{1em}
\floatbox{table}[0.645\linewidth]
{\caption{\small{Human parsing evaluation on Human3.6M, LIP and CIHP val sets with mIoU. }}
 \label{tab:parsing} }
 { \footnotesize
 \resizebox{\linewidth}{!}{
\begin{tabular}{lccc}
    \toprule
    Method & H3.6M  & LIP   & CIHP \\
    \midrule
    HCMOCO~\cite{hong2022hcmoco} & 62.50  & -     & - \\
    SNT~\cite{ji2020learning}   & -     & 54.73  & 60.87  \\
    PCNet~\cite{zhang2020part} & -     & 57.03  & 61.05  \\
    SCHP~\cite{li2020self}  & -     & 59.36  & - \\
    CDGNet~\cite{liu2022cdgnet} & -     & 60.30  & 65.56  \\
    \midrule
    UniHCP (direct eval) & 65.90  & 63.80  & 68.60  \\
    UniHCP (finetune) & \textbf{65.95}  & \textbf{63.86}  & \textbf{69.80}  \\
    \bottomrule
    \end{tabular}
 }
 }
\end{floatrow}
\end{table*}

\begin{table*}[ht]
\begin{floatrow}
\floatbox{table}[0.8\linewidth]
{\caption{\small{Pedestrian detection evaluation on CrowdHuman val set. Compared with the SOTA, UniHCP achieves comparable mAP and better JI.}}
 \label{tab:det} }
{
\footnotesize
\resizebox{\linewidth}{!}{
\begin{tabular}{lccc}
    \toprule
    Method & mAP    & MR$^{-2}(\downarrow)$  & JI \\
    \midrule
    DETR~\cite{carion2020end}  & 75.9  & 73.2  & 74.4  \\
    PEDR~\cite{lin2020detr}  & 91.6  & 43.7  & 83.3  \\
    Deformable-DETR~\cite{zhu2020deformable} & 91.5  & 43.7  & 83.1  \\
    Sparse-RCNN~\cite{sun2021sparse} & 91.3  & 44.8  & 81.3  \\
    Iter-Deformable-DETR~\cite{zheng2022progressive} & 92.1  & \textbf{41.5}  & 84.0  \\
    Iter-Sparse-RCNN~\cite{zheng2022progressive} & \textbf{92.5 }& 41.6  & 83.3  \\
    \midrule
    UniHCP (direct eval) & 90.0  & 46.6      & 82.2 \\
    UniHCP (finetune) & \textbf{92.5}  & 41.6  & \textbf{85.8} \\
    \bottomrule
    \end{tabular}
    }
    }
\hspace{-1.4em}
\floatbox{figure}[1.22\linewidth]
{\captionof{table}{\small{Pose estimation evaluation on COCO, Human3.6M, AI Challenge and OCHuman. Following~\cite{xu2022vitpose}, we report the results on COCO val set, Human3.6M, AI Challenge val set, and OCHuman test set. \dag denotes the results reported by MMPose~\cite{mmpose2020}. \ddag denotes the results achieved using multi-dataset training. }}
 \label{tab:pose}}
{ \footnotesize
\resizebox{\linewidth}{!}{
   \begin{tabular}{lcccc}
    \toprule
    Method & COCO/mAP & H3.6M/EPE($\downarrow$) & AIC/mAP 
& OCHuman/mAP \\
    \toprule
    HRNet-w32\dag~\cite{sun2019deep} &74.4 &9.4 &- &- \\
    HRNet-w48\dag~\cite{sun2019deep} & 75.1  & 7.4   & - & - \\
    TokenPose-L/D24~\cite{li2021tokenpose} & 75.9  & -     & - & -\\
    HRFormer-B~\cite{yuan2021hrformer} & 75.6  & -     & - & -\\
    ViTPose-B~\cite{xu2022vitpose} & 75.8  & -     & - & -\\
    ViTPose-B\ddag~\cite{xu2022vitpose} & \textbf{77.1}  & -     & 32.0 & 87.3 \\
    \midrule
    UniHCP (direct eval) & 76.1  & 6.9   & 32.5 & \textbf{87.4} \\
    UniHCP (finetune) & 76.5  & \textbf{6.6}   & \textbf{33.6} & N/A \\
    \bottomrule
    \end{tabular}%
    }
    }
\hspace{1.95em}
\end{floatrow}
\vspace{-0.7em}
\end{table*}

\subsection{Main Results}
To demonstrate the capability of UniHCP as a unified model for human-centric perceptions, we first evaluate our UniHCP on thirteen datasets that appear in the pretraining stage (in Section~\ref{sec:in pretrain task}), \emph{e.g.}, CIHP. Furthermore, we employ five datasets whose training splits are not included in the pretraining stage to evaluate the cross-datasets transferability of UniHCP (in Section~\ref{sec:transfer results}). We also demonstrate that UniHCP has the potential to efficiently transfer to new datasets that do not appear in pretraining with only a few images (in Section~\ref{sec:data efficiency}). For detailed evaluation configuration, please refer to the supplementary.

\subsubsection{In-pretrain Dataset Results}
\label{sec:in pretrain task}
We conduct extensive evaluations on thirteen in-pretrain datasets to demonstrate the effectiveness of our UniHCP. 
Table~\ref{tab:reid}-\ref{tab:pose} summarize the evaluation results of UniHCP on five representative human-centric tasks, \emph{i.e.}, person ReID, pedestrian attribute recognition, human parsing, pedestrian detection, and pose estimation. We report two kinds of evaluation results of our UniHCP: (1) \textbf{direct evaluation}, where the pre-trained model with cross-task shared encoder-decoder weights and task-specific queries are directly used for evaluation on the target dataset, and (2) \textbf{finetuning}, where the pretrained UniHCP are first finetuned with the train split of the target dataset and then evaluated.

As observed, the direct evaluation results of UniHCP show promising performance on most human-centric tasks, especially on human parsing and pose estimation tasks, which show better or on-par performance with the State-Of-The-Art (SOTA). The exception is the person ReID task, which observes noticeable performance gaps with the SOTA. We suggest this is due to its huge disparity from other tasks and can be remedied with quick finetuning.

With finetuning, our UniHCP achieves new SOTAs on nine out of the total twelve datasets and on par performance on the rest three datasets, even without task-specific design in architecture or task-specific priors, showing that UniHCP extracts complementary knowledge among human-centric tasks.
Concretely, Table~\ref{tab:attribute} shows that in the human attribute recognition task, UniHCP significantly surpasses previous SOTA DAFL~\cite{jia2022learning} by \textbf{+3.79\%} mA on PA-100K and \textbf{+1.20\%} mA on RAPv2 datasets, respectively, which indicates that UniHCP well extracts the shared attribute information among using the output unit of global probabilities in the interpreter. Second, UniHCP also pushes the performance of another important human task, \emph{i.e.}, human parsing, to a new level. Specifically, \textbf{+3.45\%} mIoU, \textbf{+3.56\%} mIoU, and \textbf{+4.24\%} mIoU performance gains are observed on Human3.6M, LIP, and CIHP datasets, respectively. We suggest the newly-added global supervision {$\mathcal{L}_{par}$} will help UniHCP to suppress the false prediction on not appeared semantic parts. UniHCP also shows superior performance to previous methods on pose estimation. On person ReID, UniHCP outperforms TransReid~\cite{he2021transreid} on Market1501 and MNE~\cite{li2019memory} on CUHK03 without the help of any additional camera information and training images during evaluation. For pedestrian detection, our UniHCP achieves \textbf{+1.8\%} JI performance gain compared with Iter-Deformable-DETR~\cite{zheng2022progressive} and on-par performance with the Iter-Sparse-RCNN~\cite{zheng2022progressive} on mAP. These strong performances on diverse datasets across five tasks demonstrate the feasibility and powerfulness of the unified human-centric model and large-scale pretraining.

\begin{table}[t]
\footnotesize
  \centering
  \caption{Transfer performance on ATR~\cite{liang2015human}, SenseReID~\cite{zhao2017spindle}, Caltech~\cite{dollar2011pedestrian}, MPII~\cite{andriluka20142d} and PETA~\cite{deng2014pedestrian}. Results with \dag are achieved by using additional data. DE - direct evaluation. FT - finetuning.}
  \resizebox{\linewidth}{!}{
    \begin{tabular}{lccccc}
    \toprule
    \multicolumn{1}{l}{\multirow{2}[2]*{Methods}}      & Parsing & ReID  & Detection & {Pose}  & Attribute \\
\cmidrule(r){2-2}  \cmidrule(r){3-3} \cmidrule(r){4-4} \cmidrule(r){5-5} \cmidrule(r){6-6}         & ATR   & SenseReID & Caltech($\downarrow$) & MPII   & PETA \\
    \midrule
 SOTA  & 97.39~\cite{liu2022cdgnet} & 34.6~\cite{zhao2017spindle}  & 46.6~\cite{hasan2021generalizable}  & 
 92.3~\cite{yang2021transpose}  & 87.07~\cite{jia2022learning} \\
 SOTA\dag & - & - & 28.8~\cite{hasan2021generalizable} &\textbf{93.3}~\cite{xu2022vitpose} & - \\
 \midrule
    UniHCP (DE) & -     & \textbf{46.0}    & 37.8 & -     & - \\
     UniHCP (FT)    & \textbf{97.74} & N/A     & \textbf{27.2} & 93.2  & \textbf{88.78} \\
    \bottomrule
    \end{tabular}%
  \label{tab:transfer}%
  }
  \vspace{-0.5em}
\end{table}%

\begin{table}[t]
\footnotesize
  \centering
  \footnotesize
  \caption{One-shot human parsing and human pose estimation transfer results under different tuning settings. Every method uses only 1 image per class to transfer. We repeat each experiment for 10 times and report the mean and standard deviation.}
  \resizebox{\linewidth}{!}{
    \begin{tabular}{lccc}
    \toprule
   \multicolumn{1}{l}{\multirow{2}[2]*{Methods}}  & \multicolumn{1}{l}{\multirow{2}[2]*{\makecell[c]{Learnable \\params ratio}}}    & \multicolumn{1}{c}{Parsing} & \multicolumn{1}{c}{Pose} \\
 \cmidrule(r){4-4}  \cmidrule(r){3-3}  &  & ATR/pACC    & MPII/PCKh \\
    \midrule
    One-shot finetuning & 100\%  & 90.49$\pm$1.22        & 70.6$\pm$7.53   \\
    One-shot prompt tuning &\textless{}1\% & 93.65$\pm$0.77       & 83.8$\pm$5.08   \\
    \midrule
    Full-data finetuning & 100\% & 97.74 & 93.2 \\
    \bottomrule
    \end{tabular}%
  \label{tab:prompt}%
  }
  
\end{table}%

\subsubsection{Cross-datasets Transfer Results} \label{sec:transfer results}
As the task-guided interpreter formulates all the requests of human-centric tasks into four output units, human-centric knowledge learned behind these units can be easily transferred to unseen datasets. We conduct evaluations on another five datasets which do not appear during pretraining to evaluate the transferability of UniHCP. UniHCP is finetuned to adapt to new datasets except for SenseReID, on which the performance is tested by direct evaluation.
As shown in Table~\ref{tab:transfer}, UniHCP outperforms existing SOTAs in 4 out of 5 datasets. Specifically, UniHCP achieves \textbf{+0.35\%} pACC, \textbf{+11.4\%} top-1, \textbf{-1.6\%} heavy occluded MR$^{-2}(\downarrow)$, \textbf{+0.1\%} PCKh, and \textbf{+1.71\%} mA on ATR, SenseReID, Caltech, MPII, and PETA, respectively. On MPII, UniHCP achieves on-par performance with multi-datasets trained SOTA while improving single-dataset trained SOTA by \textbf{+0.9\%} PCKh. 
Notably, even without finetuning, UniHCP achieves a \textbf{-8.8\%} heavy occluded MR$^{-2}(\downarrow)$ performance gain on single-dataset trained SOTA.
Consistent improvements on transfer tasks provide strong support
to the decent transferability of UniHCP.

\begin{table}[tbp]
  \centering
  \caption{Comparison of different parameter-sharing schemes. 
 We report the average scores of direct evaluation results on in-pretrain human-centric datasets.  ``by $\mathbf{t}_t$'' denotes sharing decoder and interpreter across task types $\mathbf{t}_t$.
For more detailed results on each dataset, please refer to the supplementary.}
  \resizebox{\textwidth}{!}{
    \begin{tabular}{lcccccc}
    \toprule
   \multicolumn{1}{c}{\multirow{2}[2]*{Methods}} & \multicolumn{1}{c}{\multirow{2}[2]{*}{\makecell[c]{Total\\ params.}}} &
 \multicolumn{1}{c}{\multirow{2}[2]{*}{\makecell[c]{Shared\\ params.}}}   & \multicolumn{3}{c}{Shared module}   & \multicolumn{1}{c}{\multirow{2}[2]*{Avg.}}\\
\cmidrule(r){4-6}  & &    & Encoder & Decoder & \makecell[c]{Task-guided\\ Interpreter}        & \\ 
\midrule
Baseline &109.32M &109.08M &  \checkmark    &   \checkmark     &   \checkmark      &67.4 \\
    (a) &156.17M &105.60M  &  \checkmark     &   \checkmark     &               & 67.4\\
    (b) &489.67M &91.07M  &  \checkmark            &       &       &60.6 \\
    (c) &170.83M &109.08M  &  \checkmark            &    by $\mathbf{t}_t$   &   by $\mathbf{t}_t$    &65.0 \\
    \bottomrule
    \end{tabular}%
    }
  \label{tab:ablation}%
  \vspace{-1em}
\end{table}%

\subsubsection{Data-Efficient Transferring} \label{sec:data efficiency}

As UniHCP achieves SOTAs on full-data finetuning setting, we further evaluate its potential for transferring to new datasets with extremely scarce training images, \emph{e.g.}, only one image per class for training. As summarized in Table~\ref{tab:prompt}, by conducting prompt tuning with one image per class, UniHCL achieves \textbf{93.65\%} pACC on ATR for parsing and \textbf{83.8\%} PCKh on MPII for pose estimation, respectively. For prompt tuning on ATR, we follow~\cite{liu2021p}. For prompt tuning on MPII, we only update queries and their associate position embeddings.
The prompt tuning results are close to that of the full-data finetuning setting and suppress the results of finetuning the whole model with one image per class for a large margin. Moreover, UniHCP with prompt tuning shows much lower standard deviations than one-shot finetuning on human parsing and pose estimation tasks, verifying that UniHCP learns generic human-centric representation which 
is beneficial for data-efficient transferring with low computation cost.

\subsection{Ablation Study on Weight Sharing}
\label{sec:ablation}
As UniHCP achieves desirable performance on various human-centric tasks while sharing most parameters among different tasks, one problem remains whether more task-specific parameters benefit learning. To answer the question, we ablate three weight sharing variants of UniHCP during pretraining using a 60k-iteration training schedule with 1k batch size. Results in Table~\ref{tab:ablation}(b) show that compared to the original UniHCP \ie, the \textit{Baseline}), unifying task-guided interpreters among all tasks resulted in an average performance on par with using specific heads while reducing about \textbf{30\%} of the parameters.  
We also note that using task-specific or task-type-specific decoders and interpreters leads to an obvious (\textbf{-6.8\%} and \textbf{-2.4\%}, respectively) performance drop on average when compared to the original UniHCP (see results in Table~\ref{tab:ablation}(b) and (c)). We speculate that in these ablation settings, complementary human-centric knowledge can not be properly shared among tasks, which leads to performance drops on most tasks.

\section{Conclusions}
\label{sec:conclu}
In this work, we present a Unified Model for Human-Centric Perceptions (UniHCP). 
Based on a simple query-based task formulation, UniHCP can easily handle multiple distinctly defined human-centric tasks simultaneously. Extensive experiments on diverse datasets demonstrate that UniHCP pretrained on a massive collection of human-centric datasets delivers a competitive performance compared with task-specific models. When adapted to specific tasks, UniHCP obtains a series of SOTA performances over a wide spectrum of human-centric benchmarks. Further analysis also demonstrate the capability of UniHCP on parameter and data-efficient transfer and the benefit of weight sharing designs. We hope our work can motivate more future works on developing general human-centric models. 

\noindent\textbf{Acknowledgement.} This paper was supported by the Australian Research Council Grant DP200103223, Australian Medical Research Future Fund MRFAI000085, CRC-P Smart Material Recovery Facility (SMRF) – Curby Soft Plastics, and CRC-P ARIA - Bionic Visual-Spatial Prosthesis for the Blind.

{\small
\bibliographystyle{ieee_fullname}
\bibliography{egbib}

\begin{thebibliography}{100}\itemsep=-1pt

\bibitem{andriluka2018posetrack}
Mykhaylo Andriluka, Umar Iqbal, Eldar Insafutdinov, Leonid Pishchulin, Anton
  Milan, Juergen Gall, and Bernt Schiele.
\newblock Posetrack: A benchmark for human pose estimation and tracking.
\newblock In {\em Proceedings of the IEEE conference on computer vision and
  pattern recognition}, pages 5167--5176, 2018.

\bibitem{andriluka20142d}
Mykhaylo Andriluka, Leonid Pishchulin, Peter Gehler, and Bernt Schiele.
\newblock 2d human pose estimation: New benchmark and state of the art
  analysis.
\newblock In {\em Proceedings of the IEEE Conference on computer Vision and
  Pattern Recognition}, pages 3686--3693, 2014.

\bibitem{aribandi2021ext5}
Vamsi Aribandi, Yi Tay, Tal Schuster, Jinfeng Rao, Huaixiu~Steven Zheng,
  Sanket~Vaibhav Mehta, Honglei Zhuang, Vinh~Q Tran, Dara Bahri, Jianmo Ni,
  et~al.
\newblock Ext5: Towards extreme multi-task scaling for transfer learning.
\newblock {\em arXiv preprint arXiv:2111.10952}, 2021.

\bibitem{FairScale2021}
FairScale authors.
\newblock Fairscale: A general purpose modular pytorch library for high
  performance and large scale training.
\newblock \url{https://github.com/facebookresearch/fairscale}, 2021.

\bibitem{braun2019eurocity}
Markus Braun, Sebastian Krebs, Fabian Flohr, and Dariu~M Gavrila.
\newblock Eurocity persons: A novel benchmark for person detection in traffic
  scenes.
\newblock {\em IEEE transactions on pattern analysis and machine intelligence},
  41(8):1844--1861, 2019.

\bibitem{cao2021handcrafted}
Jiale Cao, Yanwei Pang, Jin Xie, Fahad~Shahbaz Khan, and Ling Shao.
\newblock From handcrafted to deep features for pedestrian detection: a survey.
\newblock {\em IEEE transactions on pattern analysis and machine intelligence},
  2021.

\bibitem{carion2020end}
Nicolas Carion, Francisco Massa, Gabriel Synnaeve, Nicolas Usunier, Alexander
  Kirillov, and Sergey Zagoruyko.
\newblock End-to-end object detection with transformers.
\newblock In {\em European conference on computer vision}, pages 213--229.
  Springer, 2020.

\bibitem{chen2022unified}
Ting Chen, Saurabh Saxena, Lala Li, Tsung-Yi Lin, David~J Fleet, and Geoffrey
  Hinton.
\newblock A unified sequence interface for vision tasks.
\newblock {\em arXiv preprint arXiv:2206.07669}, 2022.

\bibitem{chen2020dynamic}
Yinpeng Chen, Xiyang Dai, Mengchen Liu, Dongdong Chen, Lu Yuan, and Zicheng
  Liu.
\newblock Dynamic convolution: Attention over convolution kernels.
\newblock In {\em Proceedings of the IEEE/CVF Conference on Computer Vision and
  Pattern Recognition}, pages 11030--11039, 2020.

\bibitem{cheng2022masked}
Bowen Cheng, Ishan Misra, Alexander~G Schwing, Alexander Kirillov, and Rohit
  Girdhar.
\newblock Masked-attention mask transformer for universal image segmentation.
\newblock In {\em Proceedings of the IEEE/CVF Conference on Computer Vision and
  Pattern Recognition}, pages 1290--1299, 2022.

\bibitem{mmpose2020}
MMPose Contributors.
\newblock Openmmlab pose estimation toolbox and benchmark.
\newblock \url{https://github.com/open-mmlab/mmpose}, 2020.

\bibitem{cust2019machine}
Emily~E Cust, Alice~J Sweeting, Kevin Ball, and Sam Robertson.
\newblock Machine and deep learning for sport-specific movement recognition: A
  systematic review of model development and performance.
\newblock {\em Journal of sports sciences}, 37(5):568--600, 2019.

\bibitem{deng2014pedestrian}
Yubin Deng, Ping Luo, Chen~Change Loy, and Xiaoou Tang.
\newblock Pedestrian attribute recognition at far distance.
\newblock In {\em Proceedings of the 22nd ACM international conference on
  Multimedia}, pages 789--792, 2014.

\bibitem{dollar2011pedestrian}
Piotr Dollar, Christian Wojek, Bernt Schiele, and Pietro Perona.
\newblock Pedestrian detection: An evaluation of the state of the art.
\newblock {\em IEEE transactions on pattern analysis and machine intelligence},
  34(4):743--761, 2011.

\bibitem{dosovitskiy2020image}
Alexey Dosovitskiy, Lucas Beyer, Alexander Kolesnikov, Dirk Weissenborn,
  Xiaohua Zhai, Thomas Unterthiner, Mostafa Dehghani, Matthias Minderer, Georg
  Heigold, Sylvain Gelly, et~al.
\newblock An image is worth 16x16 words: Transformers for image recognition at
  scale.
\newblock {\em arXiv preprint arXiv:2010.11929}, 2020.

\bibitem{fang2022alphapose}
Hao-Shu Fang, Jiefeng Li, Hongyang Tang, Chao Xu, Haoyi Zhu, Yuliang Xiu,
  Yong-Lu Li, and Cewu Lu.
\newblock Alphapose: Whole-body regional multi-person pose estimation and
  tracking in real-time.
\newblock {\em arXiv preprint arXiv:2211.03375}, 2022.

\bibitem{fei2006one}
Li Fei-Fei, Robert Fergus, and Pietro Perona.
\newblock One-shot learning of object categories.
\newblock {\em IEEE transactions on pattern analysis and machine intelligence},
  28(4):594--611, 2006.

\bibitem{ge2019deepfashion2}
Yuying Ge, Ruimao Zhang, Xiaogang Wang, Xiaoou Tang, and Ping Luo.
\newblock Deepfashion2: A versatile benchmark for detection, pose estimation,
  segmentation and re-identification of clothing images.
\newblock In {\em Proceedings of the IEEE/CVF conference on computer vision and
  pattern recognition}, pages 5337--5345, 2019.

\bibitem{gong2018instance}
Ke Gong, Xiaodan Liang, Yicheng Li, Yimin Chen, Ming Yang, and Liang Lin.
\newblock Instance-level human parsing via part grouping network.
\newblock In {\em Proceedings of the European conference on computer vision
  (ECCV)}, pages 770--785, 2018.

\bibitem{gong2017look}
Ke Gong, Xiaodan Liang, Dongyu Zhang, Xiaohui Shen, and Liang Lin.
\newblock Look into person: Self-supervised structure-sensitive learning and a
  new benchmark for human parsing.
\newblock In {\em Proceedings of the IEEE conference on computer vision and
  pattern recognition}, pages 932--940, 2017.

\bibitem{hasan2021generalizable}
Irtiza Hasan, Shengcai Liao, Jinpeng Li, Saad~Ullah Akram, and Ling Shao.
\newblock Generalizable pedestrian detection: The elephant in the room.
\newblock In {\em Proceedings of the IEEE/CVF Conference on Computer Vision and
  Pattern Recognition}, pages 11328--11337, 2021.

\bibitem{he2022masked}
Kaiming He, Xinlei Chen, Saining Xie, Yanghao Li, Piotr Doll{\'a}r, and Ross
  Girshick.
\newblock Masked autoencoders are scalable vision learners.
\newblock In {\em Proceedings of the IEEE/CVF Conference on Computer Vision and
  Pattern Recognition}, pages 16000--16009, 2022.

\bibitem{he2021transreid}
Shuting He, Hao Luo, Pichao Wang, Fan Wang, Hao Li, and Wei Jiang.
\newblock Transreid: Transformer-based object re-identification.
\newblock In {\em Proceedings of the IEEE/CVF international conference on
  computer vision}, pages 15013--15022, 2021.

\bibitem{hong2022hcmoco}
Fangzhou Hong, Liang Pan, Zhongang Cai, and Ziwei Liu.
\newblock Versatile multi-modal pre-training for human-centric perception.
\newblock {\em arXiv preprint arXiv:2203.13815}, 2022.

\bibitem{hu2021unit}
Ronghang Hu and Amanpreet Singh.
\newblock Unit: Multimodal multitask learning with a unified transformer.
\newblock In {\em Proceedings of the IEEE/CVF International Conference on
  Computer Vision}, pages 1439--1449, 2021.

\bibitem{h36m_pami}
Catalin Ionescu, Dragos Papava, Vlad Olaru, and Cristian Sminchisescu.
\newblock Human3.6m: Large scale datasets and predictive methods for 3d human
  sensing in natural environments.
\newblock {\em IEEE Transactions on Pattern Analysis and Machine Intelligence},
  36(7):1325--1339, jul 2014.

\bibitem{ji2020learning}
Ruyi Ji, Dawei Du, Libo Zhang, Longyin Wen, Yanjun Wu, Chen Zhao, Feiyue Huang,
  and Siwei Lyu.
\newblock Learning semantic neural tree for human parsing.
\newblock In {\em European Conference on Computer Vision}, pages 205--221.
  Springer, 2020.

\bibitem{jia2021spatial}
Jian Jia, Xiaotang Chen, and Kaiqi Huang.
\newblock Spatial and semantic consistency regularizations for pedestrian
  attribute recognition.
\newblock In {\em Proceedings of the IEEE/CVF international conference on
  computer vision}, pages 962--971, 2021.

\bibitem{jia2022learning}
Jian Jia, Naiyu Gao, Fei He, Xiaotang Chen, and Kaiqi Huang.
\newblock Learning disentangled attribute representations for robust pedestrian
  attribute recognition.
\newblock 2022.

\bibitem{jin2020semantics}
Xin Jin, Cuiling Lan, Wenjun Zeng, Guoqiang Wei, and Zhibo Chen.
\newblock Semantics-aligned representation learning for person
  re-identification.
\newblock In {\em Proceedings of the AAAI Conference on Artificial
  Intelligence}, volume~34, pages 11173--11180, 2020.

\bibitem{kalantidis2013getting}
Yannis Kalantidis, Lyndon Kennedy, and Li-Jia Li.
\newblock Getting the look: clothing recognition and segmentation for automatic
  product suggestions in everyday photos.
\newblock In {\em Proceedings of the 3rd ACM conference on International
  conference on multimedia retrieval}, pages 105--112, 2013.

\bibitem{kalayeh2018human}
Mahdi~M Kalayeh, Emrah Basaran, Muhittin G{\"o}kmen, Mustafa~E Kamasak, and
  Mubarak Shah.
\newblock Human semantic parsing for person re-identification.
\newblock In {\em Proceedings of the IEEE conference on computer vision and
  pattern recognition}, pages 1062--1071, 2018.

\bibitem{khamis2014joint}
Sameh Khamis, Cheng-Hao Kuo, Vivek~K Singh, Vinay~D Shet, and Larry~S Davis.
\newblock Joint learning for attribute-consistent person re-identification.
\newblock In {\em European conference on computer vision}, pages 134--146.
  Springer, 2014.

\bibitem{kokkinos2017ubernet}
Iasonas Kokkinos.
\newblock Ubernet: Training a universal convolutional neural network for low-,
  mid-, and high-level vision using diverse datasets and limited memory.
\newblock In {\em Proceedings of the IEEE conference on computer vision and
  pattern recognition}, pages 6129--6138, 2017.

\bibitem{kolesnikov2022uvim}
Alexander Kolesnikov, Andr{\'e}~Susano Pinto, Lucas Beyer, Xiaohua Zhai,
  Jeremiah Harmsen, and Neil Houlsby.
\newblock Uvim: A unified modeling approach for vision with learned guiding
  codes.
\newblock {\em arXiv preprint arXiv:2205.10337}, 2022.

\bibitem{lanchantin2020general}
Jack Lanchantin, Tianlu Wang, Vicente Ordonez, and Yanjun Qi.
\newblock General multi-label image classification with transformers.
\newblock {\em arXiv preprint arXiv:2011.14027}, 2020.

\bibitem{law2018cornernet}
Hei Law and Jia Deng.
\newblock Cornernet: Detecting objects as paired keypoints.
\newblock In {\em Proceedings of the European conference on computer vision
  (ECCV)}, pages 734--750, 2018.

\bibitem{law2019cornernet}
Hei Law, Yun Teng, Olga Russakovsky, and Jia Deng.
\newblock Cornernet-lite: Efficient keypoint based object detection.
\newblock {\em arXiv preprint arXiv:1904.08900}, 2019.

\bibitem{li2019richly}
Dangwei Li, Zhang Zhang, Xiaotang Chen, and Kaiqi Huang.
\newblock A richly annotated pedestrian dataset for person retrieval in real
  surveillance scenarios.
\newblock {\em IEEE transactions on image processing}, 28(4):1575--1590, 2019.

\bibitem{li2022mask}
Feng Li, Hao Zhang, Shilong Liu, Lei Zhang, Lionel~M Ni, Heung-Yeung Shum,
  et~al.
\newblock Mask dino: Towards a unified transformer-based framework for object
  detection and segmentation.
\newblock {\em arXiv preprint arXiv:2206.02777}, 2022.

\bibitem{li2021human}
Jiefeng Li, Siyuan Bian, Ailing Zeng, Can Wang, Bo Pang, Wentao Liu, and Cewu
  Lu.
\newblock Human pose regression with residual log-likelihood estimation.
\newblock In {\em Proceedings of the IEEE/CVF International Conference on
  Computer Vision}, pages 11025--11034, 2021.

\bibitem{li2017multiple}
Jianshu Li, Jian Zhao, Yunchao Wei, Congyan Lang, Yidong Li, Terence Sim,
  Shuicheng Yan, and Jiashi Feng.
\newblock Multiple-human parsing in the wild.
\newblock {\em arXiv preprint arXiv:1705.07206}, 2017.

\bibitem{li2020self}
Peike Li, Yunqiu Xu, Yunchao Wei, and Yi Yang.
\newblock Self-correction for human parsing.
\newblock {\em IEEE Transactions on Pattern Analysis and Machine Intelligence},
  2020.

\bibitem{li2019memory}
Suichan Li, Dapeng Chen, Bin Liu, Nenghai Yu, and Rui Zhao.
\newblock Memory-based neighbourhood embedding for visual recognition.
\newblock In {\em Proceedings of the IEEE/CVF International Conference on
  Computer Vision}, pages 6102--6111, 2019.

\bibitem{li2021uav}
Tianjiao Li, Jun Liu, Wei Zhang, Yun Ni, Wenqian Wang, and Zhiheng Li.
\newblock Uav-human: A large benchmark for human behavior understanding with
  unmanned aerial vehicles.
\newblock In {\em Proceedings of the IEEE/CVF conference on computer vision and
  pattern recognition}, pages 16266--16275, 2021.

\bibitem{li2022label2label}
Wanhua Li, Zhexuan Cao, Jianjiang Feng, Jie Zhou, and Jiwen Lu.
\newblock Label2label: A language modeling framework for multi-attribute
  learning.
\newblock In {\em European Conference on Computer Vision}, pages 562--579.
  Springer, 2022.

\bibitem{li2014deepreid}
Wei Li, Rui Zhao, Tong Xiao, and Xiaogang Wang.
\newblock Deepreid: Deep filter pairing neural network for person
  re-identification.
\newblock In {\em CVPR}, 2014.

\bibitem{li2016human}
Yining Li, Chen Huang, Chen~Change Loy, and Xiaoou Tang.
\newblock Human attribute recognition by deep hierarchical contexts.
\newblock In {\em European Conference on Computer Vision}, 2016.

\bibitem{li2022exploring}
Yanghao Li, Hanzi Mao, Ross Girshick, and Kaiming He.
\newblock Exploring plain vision transformer backbones for object detection.
\newblock {\em arXiv preprint arXiv:2203.16527}, 2022.

\bibitem{li2021tokenpose}
Yanjie Li, Shoukui Zhang, Zhicheng Wang, Sen Yang, Wankou Yang, Shu-Tao Xia,
  and Erjin Zhou.
\newblock Tokenpose: Learning keypoint tokens for human pose estimation.
\newblock In {\em Proceedings of the IEEE/CVF International Conference on
  Computer Vision}, pages 11313--11322, 2021.

\bibitem{liang2022unifying}
Jianming Liang, Guanglu Song, Biao Leng, and Yu Liu.
\newblock Unifying visual perception by dispersible points learning.
\newblock {\em arXiv preprint arXiv:2208.08630}, 2022.

\bibitem{liang2018look}
Xiaodan Liang, Ke Gong, Xiaohui Shen, and Liang Lin.
\newblock Look into person: Joint body parsing \& pose estimation network and a
  new benchmark.
\newblock {\em IEEE transactions on pattern analysis and machine intelligence},
  41(4):871--885, 2018.

\bibitem{liang2015human}
Xiaodan Liang, Chunyan Xu, Xiaohui Shen, Jianchao Yang, Si Liu, Jinhui Tang,
  Liang Lin, and Shuicheng Yan.
\newblock Human parsing with contextualized convolutional neural network.
\newblock In {\em Proceedings of the IEEE international conference on computer
  vision}, pages 1386--1394, 2015.

\bibitem{lin2018multi}
Chunze Lin, Jiwen Lu, and Jie Zhou.
\newblock Multi-grained deep feature learning for pedestrian detection.
\newblock In {\em 2018 IEEE International Conference on Multimedia and Expo
  (ICME)}, pages 1--6. IEEE, 2018.

\bibitem{lin2020detr}
Matthieu Lin, Chuming Li, Xingyuan Bu, Ming Sun, Chen Lin, Junjie Yan, Wanli
  Ouyang, and Zhidong Deng.
\newblock Detr for crowd pedestrian detection.
\newblock {\em arXiv preprint arXiv:2012.06785}, 2020.

\bibitem{lin2017focal}
Tsung-Yi Lin, Priya Goyal, Ross Girshick, Kaiming He, and Piotr Doll{\'a}r.
\newblock Focal loss for dense object detection.
\newblock In {\em Proceedings of the IEEE international conference on computer
  vision}, pages 2980--2988, 2017.

\bibitem{lin2014microsoft}
Tsung-Yi Lin, Michael Maire, Serge Belongie, James Hays, Pietro Perona, Deva
  Ramanan, Piotr Doll{\'a}r, and C~Lawrence Zitnick.
\newblock Microsoft coco: Common objects in context.
\newblock In {\em European conference on computer vision}, pages 740--755.
  Springer, 2014.

\bibitem{liu2017end}
Hao Liu, Jiashi Feng, Meibin Qi, Jianguo Jiang, and Shuicheng Yan.
\newblock End-to-end comparative attention networks for person
  re-identification.
\newblock {\em IEEE Transactions on Image Processing}, 26(7):3492--3506, 2017.

\bibitem{liu2022cdgnet}
Kunliang Liu, Ouk Choi, Jianming Wang, and Wonjun Hwang.
\newblock Cdgnet: Class distribution guided network for human parsing.
\newblock In {\em Proceedings of the IEEE/CVF Conference on Computer Vision and
  Pattern Recognition}, pages 4473--4482, 2022.

\bibitem{liu2021query2label}
Shilong Liu, Lei Zhang, Xiao Yang, Hang Su, and Jun Zhu.
\newblock Query2label: A simple transformer way to multi-label classification.
\newblock {\em arXiv preprint arXiv:2107.10834}, 2021.

\bibitem{liu2021p}
Xiao Liu, Kaixuan Ji, Yicheng Fu, Zhengxiao Du, Zhilin Yang, and Jie Tang.
\newblock P-tuning v2: Prompt tuning can be comparable to fine-tuning
  universally across scales and tasks.
\newblock {\em arXiv preprint arXiv:2110.07602}, 2021.

\bibitem{liu2017hydraplus}
Xihui Liu, Haiyu Zhao, Maoqing Tian, Lu Sheng, Jing Shao, Shuai Yi, Junjie Yan,
  and Xiaogang Wang.
\newblock Hydraplus-net: Attentive deep features for pedestrian analysis.
\newblock In {\em Proceedings of the IEEE international conference on computer
  vision}, pages 350--359, 2017.

\bibitem{loy2019wider}
Chen~Change Loy, Dahua Lin, Wanli Ouyang, Yuanjun Xiong, Shuo Yang, Qingqiu
  Huang, Dongzhan Zhou, Wei Xia, Quanquan Li, Ping Luo, et~al.
\newblock Wider face and pedestrian challenge 2018: Methods and results.
\newblock {\em arXiv preprint arXiv:1902.06854}, 2019.

\bibitem{lu2022unified}
Jiasen Lu, Christopher Clark, Rowan Zellers, Roozbeh Mottaghi, and Aniruddha
  Kembhavi.
\newblock Unified-io: A unified model for vision, language, and multi-modal
  tasks.
\newblock {\em arXiv preprint arXiv:2206.08916}, 2022.

\bibitem{lu202012}
Jiasen Lu, Vedanuj Goswami, Marcus Rohrbach, Devi Parikh, and Stefan Lee.
\newblock 12-in-1: Multi-task vision and language representation learning.
\newblock In {\em Proceedings of the IEEE/CVF Conference on Computer Vision and
  Pattern Recognition}, pages 10437--10446, 2020.

\bibitem{mao2022poseur}
Weian Mao, Yongtao Ge, Chunhua Shen, Zhi Tian, Xinlong Wang, Zhibin Wang, and
  Anton van~den Hengel.
\newblock Poseur: Direct human pose regression with transformers.
\newblock {\em arXiv preprint arXiv:2201.07412}, 2022.

\bibitem{munea2020progress}
Tewodros~Legesse Munea, Yalew~Zelalem Jembre, Halefom~Tekle Weldegebriel,
  Longbiao Chen, Chenxi Huang, and Chenhui Yang.
\newblock The progress of human pose estimation: a survey and taxonomy of
  models applied in 2d human pose estimation.
\newblock {\em IEEE Access}, 8:133330--133348, 2020.

\bibitem{nie2018mutual}
Xuecheng Nie, Jiashi Feng, and Shuicheng Yan.
\newblock Mutual learning to adapt for joint human parsing and pose estimation.
\newblock In {\em Proceedings of the European Conference on Computer Vision
  (ECCV)}, pages 502--517, 2018.

\bibitem{park2017attribute}
Seyoung Park, Bruce~Xiaohan Nie, and Song-Chun Zhu.
\newblock Attribute and-or grammar for joint parsing of human pose, parts and
  attributes.
\newblock {\em IEEE transactions on pattern analysis and machine intelligence},
  40(7):1555--1569, 2017.

\bibitem{peng2020deep}
Sida Peng, Wen Jiang, Huaijin Pi, Xiuli Li, Hujun Bao, and Xiaowei Zhou.
\newblock Deep snake for real-time instance segmentation.
\newblock In {\em Proceedings of the IEEE/CVF Conference on Computer Vision and
  Pattern Recognition}, pages 8533--8542, 2020.

\bibitem{raffel2020exploring}
Colin Raffel, Noam Shazeer, Adam Roberts, Katherine Lee, Sharan Narang, Michael
  Matena, Yanqi Zhou, Wei Li, Peter~J Liu, et~al.
\newblock Exploring the limits of transfer learning with a unified text-to-text
  transformer.
\newblock {\em J. Mach. Learn. Res.}, 21(140):1--67, 2020.

\bibitem{rezatofighi2019generalized}
Hamid Rezatofighi, Nathan Tsoi, JunYoung Gwak, Amir Sadeghian, Ian Reid, and
  Silvio Savarese.
\newblock Generalized intersection over union: A metric and a loss for bounding
  box regression.
\newblock In {\em Proceedings of the IEEE/CVF conference on computer vision and
  pattern recognition}, pages 658--666, 2019.

\bibitem{sermanet2013overfeat}
Pierre Sermanet, David Eigen, Xiang Zhang, Micha{\"e}l Mathieu, Rob Fergus, and
  Yann LeCun.
\newblock Overfeat: Integrated recognition, localization and detection using
  convolutional networks.
\newblock {\em arXiv preprint arXiv:1312.6229}, 2013.

\bibitem{shao2018crowdhuman}
Shuai Shao, Zijian Zhao, Boxun Li, Tete Xiao, Gang Yu, Xiangyu Zhang, and Jian
  Sun.
\newblock Crowdhuman: A benchmark for detecting human in a crowd.
\newblock {\em arXiv preprint arXiv:1805.00123}, 2018.

\bibitem{shazeer2018adafactor}
Noam Shazeer and Mitchell Stern.
\newblock Adafactor: Adaptive learning rates with sublinear memory cost.
\newblock In {\em International Conference on Machine Learning}, pages
  4596--4604. PMLR, 2018.

\bibitem{shu2021large}
Xiujun Shu, Xiao Wang, Xianghao Zang, Shiliang Zhang, Yuanqi Chen, Ge Li, and
  Qi Tian.
\newblock Large-scale spatio-temporal person re-identification: Algorithms and
  benchmark.
\newblock {\em IEEE Transactions on Circuits and Systems for Video Technology},
  2021.

\bibitem{su2017multi}
Chi Su, Fan Yang, Shiliang Zhang, Qi Tian, Larry~Steven Davis, and Wen Gao.
\newblock Multi-task learning with low rank attribute embedding for
  multi-camera person re-identification.
\newblock {\em IEEE transactions on pattern analysis and machine intelligence},
  40(5):1167--1181, 2017.

\bibitem{sudowe2015person}
Patrick Sudowe, Hannah Spitzer, and Bastian Leibe.
\newblock Person attribute recognition with a jointly-trained holistic cnn
  model.
\newblock In {\em Proceedings of the IEEE International Conference on Computer
  Vision Workshops}, pages 87--95, 2015.

\bibitem{sun2019deep}
Ke Sun, Bin Xiao, Dong Liu, and Jingdong Wang.
\newblock Deep high-resolution representation learning for human pose
  estimation.
\newblock In {\em Proceedings of the IEEE/CVF conference on computer vision and
  pattern recognition}, pages 5693--5703, 2019.

\bibitem{sun2021sparse}
Peize Sun, Rufeng Zhang, Yi Jiang, Tao Kong, Chenfeng Xu, Wei Zhan, Masayoshi
  Tomizuka, Lei Li, Zehuan Yuan, Changhu Wang, et~al.
\newblock Sparse r-cnn: End-to-end object detection with learnable proposals.
\newblock In {\em Proceedings of the IEEE/CVF conference on computer vision and
  pattern recognition}, pages 14454--14463, 2021.

\bibitem{sun2020circle}
Yifan Sun, Changmao Cheng, Yuhan Zhang, Chi Zhang, Liang Zheng, Zhongdao Wang,
  and Yichen Wei.
\newblock Circle loss: A unified perspective of pair similarity optimization.
\newblock In {\em Proceedings of the IEEE/CVF Conference on Computer Vision and
  Pattern Recognition}, pages 6398--6407, 2020.

\bibitem{tang2019improving}
Chufeng Tang, Lu Sheng, Zhaoxiang Zhang, and Xiaolin Hu.
\newblock Improving pedestrian attribute recognition with weakly-supervised
  multi-scale attribute-specific localization.
\newblock In {\em Proceedings of the IEEE/CVF International Conference on
  Computer Vision}, pages 4997--5006, 2019.

\bibitem{tian2015pedestrian}
Yonglong Tian, Ping Luo, Xiaogang Wang, and Xiaoou Tang.
\newblock Pedestrian detection aided by deep learning semantic tasks.
\newblock In {\em Proceedings of the IEEE conference on computer vision and
  pattern recognition}, pages 5079--5087, 2015.

\bibitem{vaswani2017attention}
Ashish Vaswani, Noam Shazeer, Niki Parmar, Jakob Uszkoreit, Llion Jones,
  Aidan~N Gomez, {\L}ukasz Kaiser, and Illia Polosukhin.
\newblock Attention is all you need.
\newblock {\em Advances in neural information processing systems}, 30, 2017.

\bibitem{vendrow2022jrdb}
Edward Vendrow, Duy~Tho Le, and Hamid Rezatofighi.
\newblock Jrdb-pose: A large-scale dataset for multi-person pose estimation and
  tracking.
\newblock {\em arXiv preprint arXiv:2210.11940}, 2022.

\bibitem{vonMarcard2018}
Timo von Marcard, Roberto Henschel, Michael Black, Bodo Rosenhahn, and Gerard
  Pons-Moll.
\newblock Recovering accurate 3d human pose in the wild using imus and a moving
  camera.
\newblock In {\em European Conference on Computer Vision (ECCV)}, sep 2018.

\bibitem{wang2020high}
Guan'an Wang, Shuo Yang, Huanyu Liu, Zhicheng Wang, Yang Yang, Shuliang Wang,
  Gang Yu, Erjin Zhou, and Jian Sun.
\newblock High-order information matters: Learning relation and topology for
  occluded person re-identification.
\newblock In {\em Proceedings of the IEEE/CVF conference on computer vision and
  pattern recognition}, pages 6449--6458, 2020.

\bibitem{wang2022ofa}
Peng Wang, An Yang, Rui Men, Junyang Lin, Shuai Bai, Zhikang Li, Jianxin Ma,
  Chang Zhou, Jingren Zhou, and Hongxia Yang.
\newblock Ofa: Unifying architectures, tasks, and modalities through a simple
  sequence-to-sequence learning framework.
\newblock In {\em International Conference on Machine Learning}, pages
  23318--23340. PMLR, 2022.

\bibitem{wang2020hierarchical}
Wenguan Wang, Hailong Zhu, Jifeng Dai, Yanwei Pang, Jianbing Shen, and Ling
  Shao.
\newblock Hierarchical human parsing with typed part-relation reasoning.
\newblock In {\em Proceedings of the IEEE/CVF conference on computer vision and
  pattern recognition}, pages 8929--8939, 2020.

\bibitem{wang2022anchor}
Yingming Wang, Xiangyu Zhang, Tong Yang, and Jian Sun.
\newblock Anchor detr: Query design for transformer-based detector.
\newblock In {\em Proceedings of the AAAI conference on artificial
  intelligence}, volume~36, pages 2567--2575, 2022.

\bibitem{wei2018person}
Longhui Wei, Shiliang Zhang, Wen Gao, and Qi Tian.
\newblock Person transfer gan to bridge domain gap for person
  re-identification.
\newblock In {\em Proceedings of the IEEE conference on computer vision and
  pattern recognition}, pages 79--88, 2018.

\bibitem{wu2019large}
Jiahong Wu, He Zheng, Bo Zhao, Yixin Li, Baoming Yan, Rui Liang, Wenjia Wang,
  Shipei Zhou, Guosen Lin, Yanwei Fu, et~al.
\newblock Large-scale datasets for going deeper in image understanding.
\newblock In {\em 2019 IEEE International Conference on Multimedia and Expo
  (ICME)}, pages 1480--1485. IEEE, 2019.

\bibitem{xiao2018simple}
Bin Xiao, Haiping Wu, and Yichen Wei.
\newblock Simple baselines for human pose estimation and tracking.
\newblock In {\em Proceedings of the European conference on computer vision
  (ECCV)}, pages 466--481, 2018.

\bibitem{xie2020polarmask}
Enze Xie, Peize Sun, Xiaoge Song, Wenhai Wang, Xuebo Liu, Ding Liang, Chunhua
  Shen, and Ping Luo.
\newblock Polarmask: Single shot instance segmentation with polar
  representation.
\newblock In {\em Proceedings of the IEEE/CVF conference on computer vision and
  pattern recognition}, pages 12193--12202, 2020.

\bibitem{xu2022fashionformer}
Shilin Xu, Xiangtai Li, Jingbo Wang, Guangliang Cheng, Yunhai Tong, and Dacheng
  Tao.
\newblock Fashionformer: A simple, effective and unified baseline for human
  fashion segmentation and recognition.
\newblock {\em arXiv preprint arXiv:2204.04654}, 2022.

\bibitem{xu2022vitpose}
Yufei Xu, Jing Zhang, Qiming Zhang, and Dacheng Tao.
\newblock Vitpose: Simple vision transformer baselines for human pose
  estimation.
\newblock {\em arXiv preprint arXiv:2204.12484}, 2022.

\bibitem{yang2019person}
Qize Yang, Ancong Wu, and Wei-Shi Zheng.
\newblock Person re-identification by contour sketch under moderate clothing
  change.
\newblock {\em IEEE transactions on pattern analysis and machine intelligence},
  43(6):2029--2046, 2019.

\bibitem{yang2021transpose}
Sen Yang, Zhibin Quan, Mu Nie, and Wankou Yang.
\newblock Transpose: Keypoint localization via transformer.
\newblock In {\em Proceedings of the IEEE/CVF International Conference on
  Computer Vision}, pages 11802--11812, 2021.

\bibitem{yuan2021florence}
Lu Yuan, Dongdong Chen, Yi-Ling Chen, Noel Codella, Xiyang Dai, Jianfeng Gao,
  Houdong Hu, Xuedong Huang, Boxin Li, Chunyuan Li, et~al.
\newblock Florence: A new foundation model for computer vision.
\newblock {\em arXiv preprint arXiv:2111.11432}, 2021.

\bibitem{yuan2021hrformer}
Yuhui Yuan, Rao Fu, Lang Huang, Weihong Lin, Chao Zhang, Xilin Chen, and
  Jingdong Wang.
\newblock Hrformer: High-resolution vision transformer for dense predict.
\newblock {\em Advances in Neural Information Processing Systems},
  34:7281--7293, 2021.

\bibitem{zhai2022scaling}
Xiaohua Zhai, Alexander Kolesnikov, Neil Houlsby, and Lucas Beyer.
\newblock Scaling vision transformers.
\newblock In {\em Proceedings of the IEEE/CVF Conference on Computer Vision and
  Pattern Recognition}, pages 12104--12113, 2022.

\bibitem{zhai2022lit}
Xiaohua Zhai, Xiao Wang, Basil Mustafa, Andreas Steiner, Daniel Keysers,
  Alexander Kolesnikov, and Lucas Beyer.
\newblock Lit: Zero-shot transfer with locked-image text tuning.
\newblock In {\em Proceedings of the IEEE/CVF Conference on Computer Vision and
  Pattern Recognition}, pages 18123--18133, 2022.

\bibitem{zhang2016far}
Shanshan Zhang, Rodrigo Benenson, Mohamed Omran, Jan Hosang, and Bernt Schiele.
\newblock How far are we from solving pedestrian detection?
\newblock In {\em Proceedings of the iEEE conference on computer vision and
  pattern recognition}, pages 1259--1267, 2016.

\bibitem{zhang2017citypersons}
Shanshan Zhang, Rodrigo Benenson, and Bernt Schiele.
\newblock Citypersons: A diverse dataset for pedestrian detection.
\newblock In {\em Proceedings of the IEEE conference on computer vision and
  pattern recognition}, pages 3213--3221, 2017.

\bibitem{zhang2019widerperson}
Shifeng Zhang, Yiliang Xie, Jun Wan, Hansheng Xia, Stan~Z Li, and Guodong Guo.
\newblock Widerperson: A diverse dataset for dense pedestrian detection in the
  wild.
\newblock {\em IEEE Transactions on Multimedia}, 22(2):380--393, 2019.

\bibitem{zhang2019pose2seg}
Song-Hai Zhang, Ruilong Li, Xin Dong, Paul Rosin, Zixi Cai, Xi Han, Dingcheng
  Yang, Haozhi Huang, and Shi-Min Hu.
\newblock Pose2seg: Detection free human instance segmentation.
\newblock In {\em Proceedings of the IEEE/CVF Conference on Computer Vision and
  Pattern Recognition}, pages 889--898, 2019.

\bibitem{zhang2013actemes}
Weiyu Zhang, Menglong Zhu, and Konstantinos~G Derpanis.
\newblock From actemes to action: A strongly-supervised representation for
  detailed action understanding.
\newblock In {\em Proceedings of the IEEE international conference on computer
  vision}, pages 2248--2255, 2013.

\bibitem{zhang2020part}
Xiaomei Zhang, Yingying Chen, Bingke Zhu, Jinqiao Wang, and Ming Tang.
\newblock Part-aware context network for human parsing.
\newblock In {\em Proceedings of the IEEE/CVF Conference on Computer Vision and
  Pattern Recognition}, pages 8971--8980, 2020.

\bibitem{zhao2017spindle}
Haiyu Zhao, Maoqing Tian, Shuyang Sun, Jing Shao, Junjie Yan, Shuai Yi,
  Xiaogang Wang, and Xiaoou Tang.
\newblock Spindle net: Person re-identification with human body region guided
  feature decomposition and fusion.
\newblock In {\em Proceedings of the IEEE conference on computer vision and
  pattern recognition}, pages 1077--1085, 2017.

\bibitem{zheng2022progressive}
Anlin Zheng, Yuang Zhang, Xiangyu Zhang, Xiaojuan Qi, and Jian Sun.
\newblock Progressive end-to-end object detection in crowded scenes.
\newblock In {\em Proceedings of the IEEE/CVF Conference on Computer Vision and
  Pattern Recognition}, pages 857--866, 2022.

\bibitem{zheng2015scalable}
Liang Zheng, Liyue Shen, Lu Tian, Shengjin Wang, Jingdong Wang, and Qi Tian.
\newblock Scalable person re-identification: A benchmark.
\newblock In {\em Proceedings of the IEEE international conference on computer
  vision}, pages 1116--1124, 2015.

\bibitem{zheng2018modanet}
Shuai Zheng, Fan Yang, M~Hadi Kiapour, and Robinson Piramuthu.
\newblock Modanet: A large-scale street fashion dataset with polygon
  annotations.
\newblock In {\em Proceedings of the 26th ACM international conference on
  Multimedia}, pages 1670--1678, 2018.

\bibitem{zheng2019joint}
Zhedong Zheng, Xiaodong Yang, Zhiding Yu, Liang Zheng, Yi Yang, and Jan Kautz.
\newblock Joint discriminative and generative learning for person
  re-identification.
\newblock In {\em proceedings of the IEEE/CVF conference on computer vision and
  pattern recognition}, pages 2138--2147, 2019.

\bibitem{zheng2017discriminatively}
Zhedong Zheng, Liang Zheng, and Yi Yang.
\newblock A discriminatively learned cnn embedding for person reidentification.
\newblock {\em ACM transactions on multimedia computing, communications, and
  applications (TOMM)}, 14(1):1--20, 2017.

\bibitem{zhou2018adaptive}
Qixian Zhou, Xiaodan Liang, Ke Gong, and Liang Lin.
\newblock Adaptive temporal encoding network for video instance-level human
  parsing.
\newblock In {\em Proceedings of the 26th ACM international conference on
  Multimedia}, pages 1527--1535, 2018.

\bibitem{zhu2017multi}
Jianqing Zhu, Shengcai Liao, Zhen Lei, and Stan~Z Li.
\newblock Multi-label convolutional neural network based pedestrian attribute
  classification.
\newblock {\em Image and Vision Computing}, 58:224--229, 2017.

\bibitem{zhu2020deformable}
Xizhou Zhu, Weijie Su, Lewei Lu, Bin Li, Xiaogang Wang, and Jifeng Dai.
\newblock Deformable detr: Deformable transformers for end-to-end object
  detection.
\newblock {\em arXiv preprint arXiv:2010.04159}, 2020.

\bibitem{zhu2022uni}
Xizhou Zhu, Jinguo Zhu, Hao Li, Xiaoshi Wu, Hongsheng Li, Xiaohua Wang, and
  Jifeng Dai.
\newblock Uni-perceiver: Pre-training unified architecture for generic
  perception for zero-shot and few-shot tasks.
\newblock In {\em Proceedings of the IEEE/CVF Conference on Computer Vision and
  Pattern Recognition}, pages 16804--16815, 2022.

\end{thebibliography}
}

\clearpage
\appendix

\section{One-shot Transfer Results}
In this section, we provide details and full results for one-shot fine-tuning and prompt tuning on human parsing and pose estimation. For each experiment, we sample ten sets of images with different random seeds; we also grid search on both iterations and learning rates until performance converges. The reported results 
 are based on the best config found for each setting. 

\noindent{\textbf{Data sampling.}} In one-shot transfer experiments, only one image per class is used for a task~\cite{fei2006one}. Table~\ref{tab:num-data-sample-one} shows the number of sampled images on one-shot transfer tasks. Note that in UniHCP, classification tasks are multi-label classification for human parsing, pose estimation, and attribute recognition, where each query performs binary classification via the global probability unit. Therefore, we also make sure the presence of cases where a class is absent is covered in our samples. Such handling avoids the query simply learning to output 1 when the corresponding class always presents within the sampled images.
On the other hand, when a class does appear in most of the images, \emph{e.g.}, all keypoint joints in pose estimation or the background class in human parsing, we are able to achieve reasonably good results without such handling, thus we do not intentionally sample ``not present'' case for keypoint joints and background class in our experiments.

\begin{table}[htbp]
  \centering
  \footnotesize
  \caption{Number of sampled images on one-shot transfer tasks. As we can easily find pose samples with all keypoint joints present in the image and do not have to consider the case where a joint is absent as explained above, we only need one sample to perform one-shot transfer on pose estimation. }
    \begin{tabular}{lcc}
    \toprule
          & \multicolumn{1}{c}{Parsing/ATR~\cite{liang2015human}} & Pose/MPII~\cite{andriluka20142d} \\
          \midrule
    Sampled images & 3 $\sim$ 4     & 1 \\
    \bottomrule
    \end{tabular}%
  \label{tab:num-data-sample-one}%
\end{table}%

\noindent{\textbf{Number of tunable parameters.}} For fine-tuning settings, all parameters are tuned. For prompt tuning on human parsing, we follow~\cite{liu2021p,zhu2022uni} and add learnable prompt tokens in decoder layers. We update queries, additional prompt tokes, and layer normalization weights. For prompt tuning on pose estimation, we only update queries and their associate position embeddings. Table.~\ref{tab:num-params} shows the number of parameters of each learnable component in prompt tuning.

\begin{table}[htbp]
  \centering
  \caption{Number of tunable parameters for prompt tuning on human parsing, pose estimation, and pedestrian attribute recognition.}
  \resizebox{\linewidth}{!}{
    \begin{tabular}{lccc}
    \toprule
          & \multicolumn{1}{l}{Parsing/ATR} & \multicolumn{1}{l}{Pose/MPII} & \multicolumn{1}{l}{Attribute/PETA} \\
          \midrule
    Query & 9216  & 8704  & 35840 \\
    Deep prompt~\cite{liu2021p,zhu2022uni} & 32256 & \multicolumn{1}{c}{-} & \multicolumn{1}{c}{-} \\
    LN~\cite{zhu2022uni}    & 16128 & \multicolumn{1}{c}{-} & \multicolumn{1}{c}{-} \\
    \midrule
    Learnable parameter ratio & 0.053\% & 0.008\% & 0.033\% \\
    \bottomrule
    \end{tabular}%
    }
  \label{tab:num-params}%
\end{table}%

\subsection{Human Parsing}
Table~\ref{tab:oneshot parsing} shows the full  one-shot results for  fine-tuning and prompt tuning on human parsing. 
\begin{table}[htbp]
  \centering
  \caption{One-shot human parsing results on ATR, evaluated by pACC. FT - finetuning, PT - prompt tuning.}
  \resizebox{\linewidth}{!}{
    \begin{tabular}{lcccccccccccc}
    \toprule
          & 1     & 2     & 3     & 4     & 5     & 6     & 7     & 8     & 9     & 10    & \multicolumn{1}{l}{avg.} & \multicolumn{1}{l}{std.} \\
          \midrule
    FT    & 91.28  & 91.21  & 90.75  & 87.90  & 91.48  & 92.14  & 89.67  & 89.36  & 90.67  & 90.48  & 90.49  & 1.22  \\
    PT    & 93.31  & 92.99  & 93.41  & 92.31  & 93.89  & 95.16  & 93.41  & 93.81  & 94.01  & 94.23  & 93.65  & 0.77  \\
    \bottomrule
    \end{tabular}%
  \label{tab:oneshot parsing}%
  }
\end{table}%

\subsection{Pose Estimation}
Table~\ref{tab:oneshot pose} shows the full one-shot results for fine-tuning and prompt tuning on pose estimation. 
\begin{table}[htbp]
  \centering
  \caption{One-shot pose estimation results on MPII, evaluated by mAP. FT - finetuning, PT - prompt tuning.}
  \resizebox{\linewidth}{!}{
    \begin{tabular}{lcccccccccccc}
    \toprule
          & 1     & 2     & 3     & 4     & 5     & 6     & 7     & 8     & 9     & 10    & \multicolumn{1}{l}{avg.} & \multicolumn{1}{l}{std.} \\
          \midrule
    FT    & 64.18  & 78.68  & 78.18  & 60.52  & 73.71  & 67.80  & 70.44  & 57.20  & 79.26  & 76.07  & 70.60  & 7.53  \\
    PT    & 87.32  & 86.13  & 87.33  & 77.44  & 85.91  & 81.16  & 88.29  & 71.97  & 87.45  & 85.29  & 83.83  & 5.08  \\
    \bottomrule
    \end{tabular}%
  \label{tab:oneshot pose}%
  }
\end{table}%

\section{Few-shot Transfer Results for Pedestrian Attribute Recognition}
In this section, we provide the few-shot transfer results for  finetuning and prompt tuning on pedestrian attribute recognition.
Different from human parsing and pose estimation datasets, the targeted downstream pedestrian attribute recognition dataset PETA~\cite{deng2014pedestrian} contains images from ten different domains. Randomly sampling only one image per class may mislead the queries to extract domain-biased representation, and we found the one-shot result is poor for both finetuning and prompt tuning under this setting. 
Therefore, we loosen the data constraint to few-shot setting to evaluate the data-efficient transfer performance on pedestrian attribute recognition.
Similar to one-shot experiments, we conduct the experiment on ten different sets of images, grid search on hyperparameters, and report results based on the best config for each setting.

\noindent{\textbf{Data sampling.}} PETA has ten different domains and 35 different attributes. For each domain, we sample images until both ``present'' and ``not present'' cases appeared at least once for each attribute; we sample multiple times and take the one with the least samples as a few-shot dataset. It takes \textbf{68 $\sim$ 75} samples to satisfy this constraint in our experiments.

\noindent{\textbf{Number of tunable parameters.}} All parameters are tuned for finetuning. For prompt tuning, we only update queries and their associate position embeddings. The number of tunable parameters in prompt tuning is shown in Table~\ref{tab:num-params}.

\noindent{\textbf{Results.}} Table~\ref{tab:oneshot attr} shows the full few-shot results for pedestrian attribute recognition; prompt tuning achieves better performance with a smaller standard deviation. 
\begin{table}[htbp]
  \centering
  \caption{Few-shot pedestrian attribute recognition results on PETA, evaluated by mA. FT - finetuning, PT - prompt tuning.}
  \resizebox{\linewidth}{!}{
    \begin{tabular}{lcccccccccccc}
    \toprule
          & 1     & 2     & 3     & 4     & 5     & 6     & 7     & 8     & 9     & 10    & \multicolumn{1}{l}{avg.} & \multicolumn{1}{l}{std.} \\
          \midrule
 FT &  59.41  & 61.03  & 59.17  & 61.73  & 59.11  & 61.30  & 59.31  & 60.46  & 60.30  & 61.38  & 60.32  & 0.96  \\
  PT &  61.71  & 61.53  & 62.41  & 62.52  & 61.19  & 63.29  & 61.58  & 61.66  & 62.94  & 63.12  & 62.20  & 0.72  \\
    \bottomrule
    \end{tabular}%
  \label{tab:oneshot attr}%
  }
\end{table}%

\section{Full Ablation Results on Weight Sharing}
In Table~\ref{tab:detailed}, we provide full results for the ablation study in Section 4.3. UniHCP achieves comparable performance with using task-specific interpreters while sharing most of the parameters among different human-centric tasks.


\begin{table*}[htbp]
  \centering
  \caption{Detailed results for different parameter-sharing methods.}
  \resizebox{\linewidth}{!}{
    \begin{tabular}{lcccccccccccccccc}
    \toprule
    \multirow{2}[2]{*}{Methods} & \multicolumn{3}{c}{Shared module} & \multicolumn{3}{c}{Parsing/mIoU} & \multicolumn{3}{c}{ReID/mAP} & Detection//mAP & \multicolumn{3}{c}{Pose/mAP} & \multicolumn{2}{c}{Attribute/mA} & \multicolumn{1}{c}{\multirow{2}[2]{*}{Average}} \\
    
\cmidrule(r){2-4}   \cmidrule(r){5-7} \cmidrule(r){8-10} \cmidrule(r){11-11} \cmidrule(r){12-14} \cmidrule(r){15-16}       & Encoder & Decoder & Task heads     & H3.6 & LIP & CIHP & Market1501 & MSMT17  & CUHK03 & CrowdHuman & COCO  & AIC  & OCHuman & PA-100K & RAPv2 &  \\
\midrule
    Baseline &  \checkmark     & \checkmark       &  \checkmark      & 64.6  & 61.9  & 64.4  & 82.1  & 59.0  & 59.9  & 80.5  & 73.5  & 29.0  & 77.0  & 81.0  & 75.3  & 67.4  \\
    (a)   &  \checkmark      &   \checkmark     &       & 65.4  & 61.6  & 64.1  & 82.7  & 59.9  & 62.1  & 82.2  & 73.5  & 27.9  & 74.9  & 81.3  & 73.0  & 67.4  \\
    (b)   &  \checkmark      &       &       & 64.2  & 59.8  & 61.1  & 76.9  & 51.3  & 51.0  & 36.2  & 71.3  & 25.6  & 69.0  & 81.9  & 78.7  & 60.6  \\
    (c)   &  \checkmark      &   by $\mathbf{t}_t$    &   by $\mathbf{t}_t$    & 64.1  & 61.6  & 63.0  & 79.4  & 54.4  & 56.3  & 68.4  & 72.7  & 26.8  & 71.3  & 82.1  & 79.8  & 65.0  \\    
\bottomrule    \end{tabular}%
}
  \label{tab:detailed}%
\end{table*}%

\section{Additional Architecture Details}
\subsection{Task-guided Interpreter}
Since the task-guided interpreter decodes each query token independently, we formulate the interpreter design by describing the generation of each output unit element $y\in\mathbf{Y}$ from query token $q\in Q^t$.

\noindent{\textbf{Feature vector unit $\mathbf{Y}_f$:}}
as the query token is already in a feature space, we do not add any additional postprocessing. we have $y_f = q, y_f \in \mathbb{R}^{C}$, where $C$ is the output dimension of the decoder.

\noindent{\textbf{Global probability unit $\mathbf{Y}_p$:}}
we apply a 1-lyr MLP (i.e. linear projector) followed by a \texttt{sigmoid} function $\sigma$, on top of query token $q$ to yield global probability $y_p \in \mathbb{R}^{1}$.

\noindent{\textbf{Local probability map unit $\mathbf{Y}_m$:}} We denoted visual tokens from the encoder as $\mathbf{F} \in \mathbb{R}^{C_e \times H / 16 \times W / 16}$, where $C_e$ denotes the output dimension of the encoder, $H\times W$ denoted the original image size and $16$ is the patch size of ViT-B. $\mathbf{F}$ is forwarded through two consecutive  deconvolution layers with hidden dimension $C_e$ to upscale the feature map to $\Tilde{\mathbf{F}} \in \mathbb{R}^{C \times H / 4 \times W / 4}$. The query token $q$ is applied with a 3-lyr MLP to get the embedding $\Tilde{q} \in \mathbb{R}^{C}$. We obtain the final probability logit map $y_m \in \mathbb{R}^{H / 4 \times W / 4}$ by calculating the dot product between $\Tilde{q}$ and $\Tilde{\mathbf{F}}$, broadcasted in the spatial dimensions. 

\noindent{\textbf{Bounding box unit $\mathbf{Y}_{bbox}$:}} Similar with~\cite{zhu2020deformable}, the query token $q$ is applied with a 3-lyr MLP to get the box offset prediction logits $\Tilde{q} = [\alpha_{\nabla cx}, \alpha_{\nabla cx}, \alpha_h, \alpha_w], \Tilde{q} \in \mathbb{R}^{4}$. With its associated anchor point $\mathcal{A}_q = [cx, cy]$, we yield the final box prediction $y_{bbox} = [\sigma(\alpha_{\nabla cx} + \sigma^{-1}(cx)), \sigma(\alpha_{\nabla cy} + \sigma^{-1}(cy)), \sigma(\alpha_{h}), \sigma(\alpha_{w})]$, where $\sigma^{-1}$ denotes the inversed \texttt{sigmoid} function.

\subsection{Positional Embedding for Encoder}
The positional embedding for the encoder is shared across tasks and is interpolated according to the spatial size of the patch projected input image. The maximum image resolution during training is $1333 \times 800$ (or $800 \times 1333$), which will then be padded to $1344 \times 800$ before patch projection (rounded up to be divisible by patch size 16). Thus, the maximum H/W dimension for images after patch projection is 84. Accordingly, we set the number of tokens for learnable positional embedding to $84\times84=7056$.
\subsection{Decoder Positional Embedding Projector}
The positional embedding projector $proj$ follows the design in~\cite{wang2022anchor}. The coordinate is first encoded by sine-cosine position encoding function~\cite{vaswani2017attention} and then projected by a simple 2-Layer MLP.

\subsection{Auxiliary Loss:}
Apart from the loss for $Q^t_L$ after $L$-th decoder block, we also add auxiliary losses to intermediate queries for pose estimation, human parsing, and pedestrian detection following the best practices in~\cite{cheng2022masked,zheng2022progressive,carion2020end}. For pose estimation and human parsing, the auxiliary loss is calculated on $Q^t_l$ for $l \in \{0, ..., L-1\}$ following~\cite{cheng2022masked}. For pedestrian detection, the auxiliary loss is calculated on $Q^t_l$ for $l \in \{1, ..., L-1\}$ following~\cite{zheng2022progressive,carion2020end}.


\subsection{Pose Estimation}
For pose estimation, we set $\lambda_{par} = 0.001$. During the inference time, when the metric requires a confidence score for keypoint filtering and NMS (e.g. mAP), we additionally multiply the global probability  prediction $y_p$ to the confidence score and lower the visibility threshold to 0.05 accordingly.

\section{Additional Training Details}
\noindent{\textbf{Loss Weight $w_\mathcal{D}$:}} for dataset $\mathcal{D}'$, its loss weight $w_\mathcal{D'}$ is calculated as follows:
\begin{equation}
\begin{aligned}
w_\mathcal{D'} = \frac{b_\mathcal{D'} w_{\mathbf{t}_\mathcal{D'}}}{\sum_{\mathcal{D} \in \mathbb{D}} b_\mathcal{D} w_{\mathbf{t}_\mathcal{D}}} ,
\end{aligned}
\end{equation}
where $b_\mathcal{D}$ denotes the batch size allocated to dataset $\mathcal{D}$ and $w_{\mathbf{t}_\mathcal{D}}$ denotes the sample weight for task type $\mathbf{t}_\mathcal{D}$. The loss weight is normalized so that it only controls the relative weight for each dataset. Samples belonging to the same task type are treated with equal importance. Since different task types have different loss functions, image input resolution, number of samples, and convergence pattern, their loss weight should be set differently. For a reasonable loss weight trade-off between tasks, we gradually add task types one at a time in a small 10k iteration joint training setup and sweep sample weights for the newly added task type. After the hyperparameter search, we set $w_{reid} = 10, w_{par} = 1\times10^{-2}, w_{seg} = 5, w_{pose} = 2\times10^3, w_{peddet} = 2$.

\noindent{\textbf{Dataset-wise Configurations:}} we provide detailed dataset-wise training configurations in Table~\ref{tab:datasheet}. In addition to these training datasets, downstream datasets are ATR~\cite{liang2015human}, SenseReID~\cite{zhao2017spindle}, Caltech~\cite{dollar2011pedestrian}, MPII~\cite{andriluka20142d} and PETA~\cite{deng2014pedestrian}.
\begin{table*}[htbp]
  \centering
  \caption{UniHCP joint training setup. \dag the batch size for pedestrian detection is reduced due to high GPU consumption.}
  \resizebox{\textwidth}{!}{
    \begin{tabular}{ccccccccc}
    \toprule
    \multirow{2}[0]{*}{Task Type} & \multirow{2}[0]{*}{\begin{tabular}[c]{@{}c@{}}Dataset\\ $\mathcal{D}$\end{tabular}} & \multirow{2}[0]{*}{\begin{tabular}[c]{@{}c@{}}Batch Size\\ $b_\mathcal{D}$\end{tabular}} & \multicolumn{1}{c}{\multirow{2}[0]{*}{\begin{tabular}[c]{@{}c@{}}Batch Size\\ per GPU\end{tabular}}} & \multicolumn{1}{c}{\multirow{2}[0]{*}{\begin{tabular}[c]{@{}c@{}}Dataset\\ Epoch\end{tabular}}} & \multicolumn{1}{c}{\multirow{2}[0]{*}{ $b_\mathcal{D} w_{\mathbf{t}_\mathcal{D}}$}} & \multirow{2}[0]{*}{\begin{tabular}[c]{@{}c@{}}GPUs\end{tabular}} & \multicolumn{1}{c}{\multirow{2}[0]{*}{\begin{tabular}[c]{@{}c@{}}Number of\\ Samples\end{tabular}}} & \multicolumn{1}{c}{\multirow{2}[0]{*}{\begin{tabular}[c]{@{}c@{}}Sample\\ Weight $w_{\mathbf{t}_\mathcal{D}}$\end{tabular}}} \\
          &       &       &       &       &       &       &       &  \\
  \toprule
    \multicolumn{1}{c}{\multirow{6}[0]{*}{\begin{tabular}[c]{@{}c@{}}Pedestrian\\ Detection\end{tabular}}} & \multicolumn{1}{l}{CrowdHuman~\cite{shao2018crowdhuman}} & \multirow{6}[0]{*}{212$^{\dag}$} & \multirow{6}[0]{*}{4} & \multirow{6}[0]{*}{130.19$^{\dag}$} & \multirow{6}[0]{*}{424} & \multirow{6}[0]{*}{53} & \multirow{6}[0]{*}{170,687} & \multirow{6}[0]{*}{2} \\
          & \multicolumn{1}{l}{EuroCity Persons~\cite{braun2019eurocity}} &       &       &       &       &       &       &  \\
          & \multicolumn{1}{l}{CityPersons~\cite{zhang2017citypersons}} &       &       &       &       &       &       &  \\
          & \multicolumn{1}{l}{WiderPerson~\cite{zhang2019widerperson}} &       &       &       &       &       &       &  \\
          & \multicolumn{1}{l}{WiderPedestrian~\cite{loy2019wider}} &       &       &       &       &       &       &  \\
          & \multicolumn{1}{l}{COCO-Person~\cite{lin2014microsoft}} &       &       &       &       &       &       &  \\  
  \hline
    \multicolumn{1}{c}{\multirow{6}[0]{*}{\begin{tabular}[c]{@{}c@{}}Person\\ ReID\end{tabular}}} & \multicolumn{1}{l}{Market-1501~\cite{zheng2015scalable}} & \multirow{3}[0]{*}{96} & \multirow{3}[0]{*}{96} & \multirow{3}[0]{*}{199.06} & \multirow{3}[0]{*}{960} & \multirow{3}[0]{*}{1} & \multirow{3}[0]{*}{50,549} & \multirow{3}[0]{*}{10} \\
          & \multicolumn{1}{l}{CUHK03~\cite{li2014deepreid}} &       &       &       &       &       &       &  \\
          & \multicolumn{1}{l}{MSMT17~\cite{wei2018person}} &       &       &       &       &       &       &  \\ \cmidrule(r){2-9}
          & \multicolumn{1}{l}{DGMarket~\cite{zheng2019joint}} & \multirow{3}[0]{*}{415} & \multirow{3}[0]{*}{415} & \multirow{3}[0]{*}{200.04} & \multirow{3}[0]{*}{4150} & \multirow{3}[0]{*}{1} & \multirow{3}[0]{*}{217,453} & \multirow{3}[0]{*}{10} \\
          & \multicolumn{1}{l}{PRCC~\cite{yang2019person}} &       &       &       &       &       &       &  \\
          & \multicolumn{1}{l}{LaST~\cite{shu2021large}} &       &       &       &       &       &       &  \\
  \hline
    \multicolumn{1}{c}{\multirow{9}[0]{*}{\begin{tabular}[c]{@{}c@{}}Pose\\ Estimation\end{tabular}}} & \multicolumn{1}{l}{COCO-Pose~\cite{lin2014microsoft}} & 286   & 286   & 200.1 & 572000 & 1     & 149,813 & 2000 \\
          & \multicolumn{1}{l}{AI Challenger~\cite{wu2019large}} & 720   & 240   & 199.46 & 1440000 & 3     & 378,352 & 2000 \\
          & \multicolumn{1}{l}{PoseTrack~\cite{andriluka2018posetrack}} & 185   & 185   & 199.55 & 3710000 & 1     & 97,174 & 2000 \\
          & \multicolumn{1}{l}{MHP~\cite{li2017multiple}} & 77    & 77    & 199.59 & 154000 & 1     & 40,437 & 2000 \\
          & \multicolumn{1}{l}{3DPW~\cite{vonMarcard2018}} & 131   & 131   & 199.98 & 262000 & 1     & 68,663 & 2000 \\
          & \multicolumn{1}{l}{UpennAction~\cite{zhang2013actemes}} & 66    & 66    & 200.66 & 132000 & 1     & 34,475 & 2000 \\
          & \multicolumn{1}{l}{JRDB-Pose~\cite{vendrow2022jrdb}} & 266   & 266   & 200.03 & 532000 & 1     & 139,385 & 2000 \\
          & \multicolumn{1}{l}{Halpe~\cite{fang2022alphapose}} & 79    & 79    & 200.69 & 158000 & 1     & 41,263 & 2000 \\
          & \multicolumn{1}{l}{Human3.6M (pose)~\cite{h36m_pami} } & 596   & 298   & 200.11 & 1192000 & 2     & 312,187 & 2000 \\
  \hline
    \multicolumn{1}{c}{\multirow{6}[0]{*}{\begin{tabular}[c]{@{}c@{}}Human\\ Parsing\end{tabular}}} & \multicolumn{1}{l}{LIP~\cite{gong2017look}} & 58    & 58    & 199.57 & 290   & 1     & 30,462 & 5 \\
          & \multicolumn{1}{l}{CIHP~\cite{gong2018instance}} & 54    & 54    & 200.14 & 270   & 1     & 28,280 & 5 \\
          & \multicolumn{1}{l}{Deep fashion~\cite{ge2019deepfashion2}} & 364   & 52    & 198.75 & 1820  & 7     & 191,961 & 5 \\
          & \multicolumn{1}{l}{VIP~\cite{zhou2018adaptive}} & 35    & 35    & 198.63 & 175   & 1     & 18,469 & 5 \\
          & \multicolumn{1}{l}{ModaNet~\cite{zheng2018modanet}} & 100   & 50    & 200.62 & 500   & 2     & 52,245 & 5 \\
           & \multicolumn{1}{l}{Human3.6M (parse)~\cite{h36m_pami}} & 120   & 40    & 200.71 & 600   & 3     & 62,668 & 5\\
  \hline
    \multicolumn{1}{c}{\multirow{6}[0]{*}{\begin{tabular}[c]{@{}c@{}c@{}}Pedestrian\\ Attribute\\ Recognition\end{tabular}}} & \multicolumn{1}{l}{PA-100K~\cite{liu2017hydraplus}} & 172   & 172   & 200.32 & 1.72  & 1     & 90,000 & 0.01 \\
          & \multicolumn{1}{l}{RAPv2~\cite{li2019richly}} & 130   & 130   & 200.55 & 1.3   & 1     & 67,943 & 0.01 \\
          & \multicolumn{1}{l}{HARDHC~\cite{li2016human}} & 54    & 54    & 199.75 & 0.54  & 1     & 28,336 & 0.01 \\
          & \multicolumn{1}{l}{UAV-Human~\cite{li2021uav}} & 31    & 31    & 200.78 & 0.31  & 1     & 16,183 & 0.01 \\
          & \multicolumn{1}{l}{Parse27k~\cite{sudowe2015person}} & 52    & 52    & 198.33 & 0.52  & 1     & 27,482 & 0.01 \\
          & \multicolumn{1}{l}{Market-1501 (attribute)~\cite{zheng2015scalable}} & 25    & 25    & 202.57 & 0.25  & 1     & 12,936 & 0.01 \\
  \hline
    \multirow{2}[0]{*}{Summary} & \multirow{2}[0]{*}{/} & \multirow{2}[0]{*}{Total: 4324} & \multirow{2}[0]{*}{/} & \multirow{2}[0]{*}{\begin{tabular}[c]{@{}c@{}}Avg.: 200.00\\ (excluding det.)\end{tabular}} & \multirow{2}[0]{*}{/} & \multirow{2}[0]{*}{Total: 88} & \multirow{2}[0]{*}{Total: 2,327,403} & \multirow{2}[0]{*}{/} \\
          &       &       &       &       &       &       &       &  \\
    \bottomrule
    \end{tabular}%
}
  \label{tab:datasheet}%
\end{table*}%
\section{Additional Finetuning Details}
We provide major finetuning configurations in Table~\ref{tab:ftconfig}; other settings are identical to the training config.



\begin{table*}[htbp]
  \centering
  \caption{Detailed finetuning configs for human-centric tasks.}
  \resizebox{0.99\linewidth}{!}{
    \begin{tabular}{ccccccccc}
    \toprule
    {Task Type} & Dataset &{Learning Rate} & {Batch Size} & Iterations  &{Backbone lr Multiplier} & {Drop Path Rate} & {Layer Decay Rate} & {Weight Decay} \\
    \midrule
    \multicolumn{1}{c}{\multirow{2}[0]{*}{
    \makecell[c]{Pedestrian \\Detection}
    }} & CrowdHuman~\cite{shao2018crowdhuman} & 2.00E-04 & 32    & 160k  & 1.0     & 0.2   & 0.75  & 0.05 \\
          & Caltech~\cite{dollar2011pedestrian} & 1.00E-05 & 32    & 30k   & 0.1   & 0.2   & 0.75  & 0.05 \\
    \midrule
        \multirow{3}[0]{*}{\makecell[c]{Person\\ ReID}} & Market-1501~\cite{zheng2015scalable} & 1.00E-04 & 64    & 40k   & 0.4   & 0.05  & 0.75  & 0.5 \\
        & CUHK03~\cite{li2014deepreid} & 5.00E-05 & 64    & 20k   & 0.9   & 0.1   & 0.95  & 0.5 \\
          & MSMT17~\cite{wei2018person}  & 1.00E-04 & 64    & 40k   & 0.9   & 0.05  & 0.75  & 0.5 \\
          
    \midrule
    \multirow{4}[0]{*}{\makecell[c]{Pose\\ Estimation}} & COCO-Pose~\cite{lin2014microsoft}  & 1.00E-04 & 512   & 20k   & 0.9   & 0.25  & 0.75  & 0.05 \\
          
          & AI Challenger~\cite{wu2019large}   & 1.00E-03 & 512   & 10k   & 0.9   & 0.2   & 0.75  & 0.05 \\
          & Human3.6M (Pose)~\cite{h36m_pami} & 5.00E-06 & 512   & 10k   & 0.9   & 0.3   & 0.75  & 0.05 \\
          & MPII~\cite{andriluka20142d}  & 7.00E-05 & 512   & 7.5k  & 0.9   & 0.3   & 0.75  & 0.05 \\
    \midrule
    \multicolumn{1}{c}{\multirow{4}[0]{*}{\makecell[c]{Human \\Parsing}}} 
          & LIP~\cite{gong2017look}   & 5.00E-05 & 64    & 30k   & 1.0     & 0.3   & 0.75  & 0.05 \\
          & CIHP~\cite{gong2018instance}  & 1.00E-04 & 64    & 35k   & 1.4   & 0.3   & 0.65  & 0.05 \\
          & Human3.6M (parse)~\cite{h36m_pami} & 1.00E-05 & 64    & 25k   & 1.3   & 0.3   & 0.85  & 0.05 \\
          & ATR~\cite{liang2015human}   & 1.00E-04 & 64    & 15k   & 0.7   & 0.3   & 0.85  & 0.05 \\
          \midrule
    \multicolumn{1}{c}{\multirow{3}[0]{*}{\makecell[c]{Pedestrian \\Attribute\\ Recognition}}} & PA-100K~\cite{liu2017hydraplus} & 3.00E-03 & 128   & 10k   & 0.05  & 0.2   & 0.85  & 0.05 \\
          & RAPv2~\cite{li2019richly} & 5.00E-04 & 128   & 4k    & 0.5   & 0.3   & 0.75  & 0.05 \\
          & PETA~\cite{deng2014pedestrian}  & 1.00E-03 & 128   & 20k   & 0.2   & 0.3   & 0.75  & 0.05 \\
    
    \bottomrule
    \end{tabular}%
    }
  \label{tab:ftconfig}%
\end{table*}%

\section{Ethics}
In this work, we proposed a model to unify multiple human-centric tasks and trained the model on a huge collection of public and widely used human-centric datasets. We acknowledge that the resulting model demonstrates good performance on public ReID benchmarks and thus may be associated with potential identity information leaking without consent if misused. Therefore, the pretrained model will be released only on a case-by-case basis, and the requester must sign an agreement limiting the usage to research purposes only. In addition, the pretrained query tokens for ReID tasks will be excluded from the model release.

\end{document}